# What fifty-one years of Linguistics and Artificial Intelligence research tell us about their correlation: A scientometric analysis

Mohammed Q. Shormani, Ibb University, Yemen
shormani@ibbuniv.edu.ye/ https://orcid.org/0000-0002-0138-4793

**Abstract**

There is a strong correlation between linguistics and artificial intelligence (AI), best manifested by deep learning language models. This study provides a thorough scientometric analysis of this correlation, synthesizing the intellectual production over 51 years, from 1974 to 2024. Web of Science Core Collection (WoSCC) database was the data source. The data collected were analyzed by two powerful software, viz., CiteSpace and VOSviewer, through which mapping visualizations of the intellectual landscape, trending issues and (re)emerging hotspots were generated. The results indicate that in the 1980s and 1990s, linguistics and AI (AIL) research was not robust, characterized by unstable publication over time. It has, however, witnessed a remarkable increase of publication since then, reaching 1478 articles in 2023, and 546 articles in January-March timespan in 2024, involving emerging issues including *Natural language processing*, *Cross-sectional study*, *Using bidirectional encoder representation*, and *Using ChatGPT* and hotspots such as *Novice programmer, Prioritization*, and *Artificial intelligence*, addressing new horizons, new topics, and launching new applications and powerful deep learning language models including ChatGPT. It concludes that linguistics and AI correlation is established at several levels, research centers, journals, and countries shaping AIL knowledge production and reshaping its future frontiers.

**Keywords**: Linguistics, artificial intelligence, correlation, scientometrics, knowledge visualizations, trending issues, hotspots

## Introduction

The correlation between Linguistics and Artificial Intelligence (AI) appears to be largely overlooked by the scholarly community. Despite the significant correlation between linguistics and AI (AIL) in various areas (Gulordava et al. 2018; Linzen and Baroni 2021; McShane and Nirenburg 2021; Maruthi et al. 2021; Medium 2023; Shormani 2024a/d; Wang and Wen 2025), this correlation remains underrecognized, as evidenced by the limited number of publications on the topic. Considering this problem, this study aims to uncover this correlation, synthesizing AIL research over 51 years from 1974 to 2024. It provides a thorough scientometric analysis of AIL knowledge production, demarcating its landscape, trending themes, and emerging hotspots. Two powerful software, viz., CiteSpace and VOSviewer, were used to visualize and generate knowledge mappings, uncovering several types of analyses including document co-citation analysis (DCA), author co-citation analysis (ACA), keyword co-occurrence analysis (KCA), citations counts, clusters, burstness, betweenness centrality, Modularity Q, Silhouette and Sigma for which CiteSpace was used. Additionally, VOSviewer was used to generate knowledge visualizations of author's keyword(s) co-occurrence, key knowledge producers including journals, research centers/institutions, and countries. This study, thus, brings the correlation between linguistics and artificial intelligence to light, emphasizing the linguistic bases underlying AI industry. It is, to the best of our knowledge, the first study to tackle this crucial topic, specifically employing scientometric analysis.

Linguistics is the scientific study of how language evolves, how it is acquired, perceived, computed, represented, and studied in its several modules including phonology, morphology, syntax, semantics, and language acquisition (Shormani 2024a/b). On the other hand, AI refers to creating, modeling and/or producing (machine) intelligence similar to that of human. It is the development of computer models or machines that can perform tasks like human intelligence. These machine/computer models involve algorithms trained on large datasets to learn patterns and make predictions. These algorithms or artificial neural networks (ANNs) can learn deeply, and be trained with multiple layers to perform complex tasks (McShane and Nirenburg 2021). AI aims to simulate intelligent behavior including learning, problem-solving, perception, and even decision-making (Liao et al. 2018). These models are reported to perform tasks with considerable accuracy (Gulordava et al. 2018; Linzen and Baroni 2021).

Artificial intelligence was "born" in 1950, perhaps with Turing's (1950) famous question "Can Machine Think?" Two influential papers by Turing (1950) and Minsky (1961) shaped the field of AI. As early as the late 1960s and well in the early 1970s, there has arisen a strong tendency to scrutinize the relationship between AI and linguistics (Rosenberg 1975). The tendency consists in constructing computer programs that parallel human intelligence. The aim was to "understand what intelligence is and how it can be put in computers" and no doubt that "[l]anguage is one of the most complex and unique of human activities, and understanding its structure may lead to a better theory of how our minds work" (Winograd 1971: 15). In fact, the relationship between linguistics and AI is manifested from the very beginning of AI inception. One such manifestation results in shaping a branch of linguistics called Computation Linguistics (CL), a field comprising any work involving natural language processing (NLP); which has been developed since its inception in 1940s. NLP is a field of computer science and technology, the main aim of which is to make computers generate, process and interpret human language. The ideas and projects by CL/NLP and AI scientists were first crystalized in question-answering systems, machine translation, and man-machine conversation (Rosenberg 1975; Kenny 2022; Shormani 2024c).

The relationship between linguistics and AI could be seen as some sort of correlation; linguistics, with its profound analysis of phonology, syntax, and semantics, psychology, biology, provides AI with the theoretical foundations necessary for programming, training and working of language models (Medium 2023). The correlation between linguistics and AI is more than simply a juxtaposition of a linguistic research area with technology. It is rather an integrative phenomenon that unveils and enhances our understanding of how human language and AI interact and the kind of prowess they come up with. Language models are computer programs that are trained on language data to generate and process human language data. This intersection between linguistics and AI is now constituting a trending theme of linguistics-AI interaction, where language, the unprecedented human characteristic, and the prowess of AI are embraced (Medium 2023; Shormani 2024a). Thus, on one extreme, linguistics provides an invaluable procedure for AI to generate, process and/or interpret human language data.

On the other extreme, AI makes available new frameworks for research, uncovering new approaches, methods and tools for linguistics and linguistic inquiry.

The article is organized as follows. Section 2 outlines the linguistic bases underlying AI. It also reviews some related studies, focusing on their questions and findings. Section 3 spells out the study design and data analysis, addressing how the data were collected, the preference of Web of Science Core Collection (WoSCC) as the data source and outlining its limitations. This section also articulates the methods of analysis, highlighting the reasons of using CiteSpace and VOSviewer software. Section 4 presents the study results, in which we tabulate and analyze the findings reached. Section 5 discusses these results, detailing the AIL landscape, emerging trends and hotspots. In this section, we also discuss and elaborate on the findings reached, their insights into the correlation between linguistics and AI, and how this correlation results in fast and vast developments of AI-driven tools, models and applications. Section 6 concludes the article, elaborating on the study findings and providing some limitations and further implications for future research.

## 2. Theoretical foundations

### 2.1. literature review

To understand how linguistic bases underly AI, it is crucial to answer two major questions: i) What are the linguistic bases in AI? And ii) How does AI apply these linguistic bases in its working mechanisms? we tackle these questions each in turn.

As for the first question, it has been, in fact, asked by a number of scholars. For example, McShane and Nirenburg (2021) asked a similar question: What is linguistics for the age of AI? They provide a valuable answer to this question, consisting of four parts: First, it is the study of linguistics in service of developing natural language understanding and generation capabilities within an integrated, comprehensive agent architecture (p. 20). Second, it is the study of linguistics in service of developing natural language understanding and generation capabilities (1) within an integrated, comprehensive agent architecture, (2) *using human inspired, explanatory modeling techniques and actionability judgments* (p. 22, emphasis in the original). Third, it is the study of linguistics in service of developing natural language understanding and generation capabilities (1) within an integrated, comprehensive agent architecture, (2) using human-inspired, explanatory modeling techniques, and (3) *leveraging insights from linguistic scholarship and, in turn, contributing to that scholarship* (p. 34). Fourth, it is the study of linguistics in service of developing natural language understanding and generation capabilities (1) within an integrated, comprehensive agent architecture, (2) using human-inspired, explanatory modeling techniques, (3) leveraging insights from linguistic scholarship and, in turn, contributing to that scholarship, and (4) *incorporating all available heuristic evidence when extracting and representing the meaning of language inputs* (p. 40).

The answer to this question leads to scientific revolution, resulting in remarkable achievements in AIL industry- the incorporation of neural algorithms and deep neural networks (DNNs), first occurring as ANNs, into several and various technology applications including NLP, machine translation and reading comprehension (Edunov et al. 2018; Linzen and Baroni 2021). DNNs "are mathematical objects that compute functions from one sequence of real numbers to another sequence", by means of "neurons". Linguistically, DNNs "learn to encode words and sentences as vectors (sequences of real numbers); these vectors, which do not bear a transparent relationship to classic linguistic structures, are then transformed through a series of simple arithmetic operations to produce the network's output" (Linzen and Baroni 2021: 196). DNNs dominate NLP and CL works "deriving semantic representations from word co-occurrence statistics" (Pavlick 2022: 447). There are also other types of deep earning networks, viz., recurrent neural networks (RNNs). These networks constitute a mechanism that encodes word sequences, in a left-to-right fashion, "maintaining a single vector, the so-called hidden state, which represents the first *t* words of the sentence" (Linzen and Baroni 2021: 197).

As for the second question, there are several studies shedding light on linguistics involvement in AI. In what follows, we will tackle and exemplify the syntactic and semantic phenomena involved in training AI learning models. Concerning syntax, there are several syntactic phenomena on which DNNs were trained such as filler-gap dependencies. Gulordava et al. (2018) have conducted an empirical study in which they train these AI learning models on how to identify the next n-gram, regardless of specifically supervising this construction. They have also trained them on a filler–gap dependency. This syntactic phenomenon could be defined as removing an NP constituent if a wh-licensor is used, thus leaving a gap, and what the AI model has to learn is to predict a gap-one of the NPs in the embedded clause (cf. Gulordava et al. 2018: 200; Shormani 2024a):

(1) a. I know that you met your brother yesterday. (no wh-licensor, no gap)
b. *I know who you met your brother yesterday. (wh-licensor, no gap)

In examples (1), (1a) is syntactically well-formed while (1b) is not, and its ungrammaticality lies in the fact that using *who* entails the omission of the NP *your brother.*

Long-distance agreement (LDA) is another syntactic phenomenon in which a constituent α agrees with a constituent *β*, where α and *β* are far from each other as in (2b) (cf. also Shormani 2024a):

(2) a. In our class, the hardworking student is Ali.

b. In our class, the hardworking student *who liked syntax books* is/*are Ali.

In (2a), the NP *the hardworking student* agrees with the verb *is*, and they are adjacent (not far from each other). However, though they are not adjacent in (2b), they agree also in all phi-features (person, number and gender). The verb *is* agrees with the subject *the hardworking student* though there are five words, or otherwise the embedded clause, *who wrote several syntax books*, between both constituents. Here the words *who liked syntax books* between the head of NP *the hardworking student*, which is *students*, and the verb *is*, are called attractors because they intervene between the subject *linguists* and the verb *is* (see also Shormani 2024a).

In the literature, LDA has received much research in syntactic inquiry across languages (see e.g. Polinsky and Potsdam 2001; Chomsky 2001, 2005, 2008; Ackema et al. 2006; Koeneman and Zeijlstra 2014; Rouveret 2008; Shormani 2017, 2024a/b). It has also received much interest in AI, specifically deep learning language models (Linzen et al. 2016; Gulordava et al. 2018; Linzen and Baroni 2021; Thrush et al. 2020). In these studies, deep learning models such as DNNs, RNNs were trained on data involving LDA, and the performance of these deep learning models was considerably high, scoring high levels of accuracy, and sometimes even surpasses humans (Gulordava et al. 2018; Kung et al. 2023). In Gulordava et al.'s (2018) experiment, for example, the accuracy rate was 82%. However, the accuracy rate of DNNs changes the more attractors we introduce. For example, DNNs were unable to predict LDA beyond 5-grams.

As for semantics, several studies have tackled the semantic-AI interaction, accelerating the semantic bases in AI and how linguistics, in general, contributes to the advancement of AI. The semantic bases in AI have been tackled in relation to several semantic phenomena. For example, Ettinger (2020) has studied how AI language models can be trained on argument structure, a semantic structure involving thematic roles such as agent and patient. In particular, Ettinger tested whether Bidirectional Encoder Representations from Transformers (BERT) can identify the argument structure and the semantic role an NP can carry in a particular sentence, differentiating between, for instance, agent and patient. Table 1 showcases an example of the argument structure data that NNAs have been trained on (Ettinger 2020, see also Shormani 2024a):

**Table 1: Examples of argument structure (from Ettinger 2020: 38)**

| **Context 1** | **Compl** | **Context 2** | **Match** | **Mismatch** |
|---|---|---|---|---|
| *The restaurant owner forgot which customer the waitress had ____* | *served* | *A robin is a _____* | *bird* | *tree* |
| *The restaurant owner forgot which waitress the customer had____* | *served* | *A robin is not a____* | *bird* | *tree* |

Ettinger (2020) utilized psycholinguistic stimuli to enhance the training process, and the performance of BERT in this experiment was good enough, i.e. 86% accuracy. Moreover, commonsense knowledge, as a semantic phenomenon, utilized in the training of neural models, was also examined by Ettinger (2020). Ettinger trained BERT to recognize hyponym–hypernym relations. The prompts used include *A robin is a [MASK]*, and BERT's performance was considerably high. For example, deciding whether *bird* or *tree* (Table 1), BERT performance was 100%. Additionally, Li et al. (2021) conducted a study, uncovering the implicit representations of meaning in neural language models. They found that dynamic representations of meaning and implicit simulation support prediction in pre-trained neural language models. The ability of BERT to identify novel verb was examined by Thrush et al. (2020). They selected a subclass of verbs based on their selectional restrictions and subcategorization restrictions and trained BERT to do certain tasks.

Neural network models have also been trained on several other semantic phenomena including compositionality, systematicity, and compositionality of negation. As for compositionality, Everaert et al. (2015: 731) state that it is a property of human language, constraining "the relation between form and meaning". It refers to the idea that the meaning of a sentence is composed of the meaning of words involved plus the pragmatic context in which this sentence is used. Concerning systematicity, it could be defined as "the ability to produce/understand some utterances is intrinsically connected to the ability to produce/understand certain others" (Fodor and Pylyshyn 1988: 37). Thus, if an ANNs model can understand the sentence: *Ahmed respects Ali*, it is expected that it understands the sentence *Ali respects Ahmed*. NNAs have also been trained on compositionality of negation, which is another semantic notion, and part of human language. According to Everaert et al. (2018), the performance of NNAs was high after being trained on sufficient data. There are also several other semantic phenomena that ANNs have been trained on including phrase representations (Shwartz and Dagan 2019), polysemy and composition (Mandera et al. 2017), among many others.

Furthermore, AI neural models can also learn interface phenomena if they have been trained on sufficient amounts of data. These issues include syntax-semantics interface (Baroni and Lenci 2010), morphology-semantics interface (Marelli and Baroni 2015), among others (cf. also Shormani and Qarabesh 2018). Thus, the competence acquired by any language model in any syntactic or semantic phenomenon is due to being trained on massive amounts of data of this phenomenon. The way these models learn/acquire a linguistic phenomenon is similar to a great extent to the way in which humans acquire a language. In the same spirit, both these neural models and humans learn/acquire any linguistic phenomenon, again if they are exposed to sufficient and efficient linguistic input necessary for language acquisition to take place (Chomsky 1981, 1995; Shormani 2014a/b, 2016, 2023). When humans acquire language, be it L1, L2, Ln, there is also much involvement of generic architectural properties and features in the same way hierarchical structures including Tree markers or Phrase markers represent how a piece of human language is derived and computed, and the mental properties and capacities involved in processing it (Chomsky 1957, 1965, 2013; Shormani 2013, 2017, 2024a/b). All these aspects are applied in NNAs' working mechanisms.

There are also recent and current studies that reflect the linguistics and generative AI intersection, tackling trending issues of NLP, ChatGPT, and AI in general. For example, Sohail et al. (2018) present a novel book recommendation system that integrates fuzzy linguistic quantifiers with Ordered Weighted Aggregation (OWA) operators. The system uses Positional Aggregation-based Scoring (PAS) to convert university book rankings into scores, applying OWA operators to customize recommendations. The evaluation, using metrics like Precision at 10 and Mean Precision, shows that the approach they propose performs comparably or even better than existing recommendation methods, emphasizing the effectiveness of fuzzy linguistic quantifiers in recommender systems. Another important study has been conducted by Maruthi et al. (2021) who critically examine the semantic frameworks underpinning human-centric AI through linguistic lens. They analyze how language and semantics influence AI system development, particularly in applications designed for human-AI collaboration. The study highlights the importance of integrating linguistic principles into AI systems to enhance their understanding of human language, addressing challenges like ambiguity and context. A further significant study was conducted by Anas et al. (2024) who aim to evaluate the effectiveness of generative AI models such as ChatGPT and Bard, in sentiment analysis compared to traditional deep learning algorithms like RoBERTa. Their experiments reveal that RoBERTa outperforms both ChatGPT and Bard, suggesting that generative AI models currently struggle to capture the nuances and subtleties of sentiment in text. These findings provide valuable insights into the strengths and weaknesses of different models for sentiment analysis, guiding researchers and practitioners in selecting appropriate models for their tasks. Additionally, Nadeem et al. (2024) investigate the integration of LLMs with deep learning techniques to enhance emotion recognition from images. While LLMs like GPT-3.5 and GPT-4 are adept at processing text, their capability to analyze visual data is limited. To bridge this gap, the authors propose combining these models with vision transformers, enabling the system to process and interpret visual inputs effectively. This hybrid approach aims to improve the accuracy and depth of emotion recognition systems by leveraging the strengths of both language and visual processing models.

Furthermore, Qamar et al. (2024) evaluate ChatGPT's ability to process linguistic ambiguity, creativity, and language games. They find that while ChatGPT can generate human-like text, it struggles with nuanced and complex language constructs, often leading to inaccuracies and misunderstandings. This study emphasizes the limitations of AI in fully grasping the subtleties of human language, especially in creative contexts. Anwar et al. (2023) propose an AI framework for resolving tied events in sports surveillance. By analyzing diverse performance metrics beyond traditional scoring systems, including player statistics and real-time dynamics, their system objectively recommends a winner in tied situations. This approach enhances decision-making in sports, providing more accurate and unbiased outcomes in surveillance contexts. A final study that could be reviewed here is done by AlSagri and Sohail (2024) who explore fractal-inspired techniques in sentiment analysis, comparing the performance of ChatGPT and Gemini Bard with traditional deep learning models. They suggest that integrating fractal mathematics with natural language processing can improve sentiment classification accuracy and contextual understanding, pushing for innovative methodologies to advance sentiment analysis. Shormani (2025) examines the ability of advanced non-native English speakers (NNSs) versus ChatGPT in processing center-embedding English sentences. Recruiting a NNS participant group of 15 individuals, the study compares human processing/interpreting of complex syntactic structures to that of ChatGPT, revealing that despite the syntactic complexity, humans consistently outperformed the AI model even in the case of NNSs. This suggests that non-native speakers, while facing challenges, still exhibit superior linguistic processing compared to ChatGPT, reinforcing the unique cognitive mechanisms underlying human language cognitive abilities. The findings highlight AI's limitations in handling deeply nested syntactic structures, emphasizing the persistent gap between human cognition and machine-based natural language processing. The study concludes that human brain is still superior to ChatGPT and that there is no difference between NNSs and native speakers of a L2 provided that these NNSs have internalized and built a "perfect" linguistic system of L2.

However, the use of AI often overlaps with ethical issues, biases and inaccuracies. This very crucial issue has been addressed by ample studies and authors (see e.g. Amer 2022; Biswas 2023; J Lee 2023; Dergaa et al. 2023; Ortega-Bolaños et al. 2024; Ortega-Bolaños et al. 2024). A recent study has been conducted by AlSagri et al. (2024) which addresses ethical concerns raised about the uncritical use of AI-driven tools in research and academic spheres, with the authors urging human oversight. They compare the performance of ChatGPT-3.5 and Gemini (Bard) in assisting scientific writing, finding that Gemini performs better with a perfect score (100%) compared to ChatGPT's score of 70%. They stress that while AI models can aid in tasks like language editing and reference formatting, relying on these models for scientific article writing is problematic due to potential inaccuracies.

Thus, given this study scope and purpose, the following research questions are specifically addressed:

1. How does the actual knowledge landscape of AIL research look like in 1974-2024 timeframe in terms of scientometric indicators including DCA, Modularity Q, Silhouette, burstness, and betweenness centrality?
2. What are the (re)emerging issues and hotspots of AIL research in 2018-2024 timeframe in terms of scientometric indicators?
3. Who are the key contributors to AIL research, and what are the possible knowledge gaps in AIL research globally?

## 4. Study design

### 4.1. Data collection

Web of Sicence (WoS) WoSCC databases were selected as the source of data to retrieve the study data from, because it is a reliable source (Liu 2019), and has a huge coverage. As of 2019, WoSCC contains "multidisciplinary information from over 18,000 high impact journals, over 180,000 conference proceedings, and over 80,000 books from around the world. With over 100 years of comprehensive coverage and more than one bi-cited reference connections" (Liu 2019: 1817, see also Asubiaro et al. 2024; Liu et al., 2024). The data were collected on March 2, 2024, in one day and one session to avoid the possible daily addition of articles to WoSCC. For scientometric review studies, WoS provides reliable sources having all the information needed to allow for an in-depth metric analysis in terms of DCA, ACA, KCA, institutions, authors, countries, among others, which software like CiteSpace and VOSviewer are fed with. The WoSCC sub-datasets and coverage years are as follows:

1. Science Citation Index Expanded (SCI-EXPANDED): 1970-present
2. Social Sciences Citation Index (SSCI): 1970-present
3. Arts & Humanities Citation Index (AHCI): 1975-present
4. Emerging Sources Citation Index (ESCI)- 2020-present

Although WoSCC has been expanded noticeably in its (sub-)databases, some of these sometimes face some sort of declension or stagnation (see Liu et al. 2024, for a comprehensive discussion).

### 4.2. Search terms

The timespan in WoS was set to 1974-2024. The year 1974 was selected as the starting point of the analysis, as no relevant publications were found prior to this date. In our data, Sobel (1974) was recorded as the first and only publication in 1974. Given the study scope and purpose, the following search terms were employed in WoS search engine "Artificial intelligence AND linguistics" OR "Natural language processing AND linguistics" OR "Linguistic bases AND artificial intelligence" OR "Semantics AND artificial intelligence" OR "Syntax AND artificial intelligence" OR "Morphology AND artificial intelligence" OR "Artificial intelligence AND language study" OR "Linguistics AND ChatGPT". These search terms resulted in 6977 articles. However, CiteSpace Remove Duplicates function identified 5750 as unique records, and 1227 were duplicated articles. To ensure unambiguous results, acronyms like "AI" and "NLP" were excluded since these forms are in fact contained in their full forms. Terms such as "Data mining", "Internet of things", "Deep learning", etc. were also excluded because the focus is actually on phrases that represent the study variables both "Linguistics" and "Artificial Intelligence" and not terms belonging either to linguistics or AI alone, and that is why Boolean operators "AND" and "OR" were used in our search strategy. When the results of each search process were obtained, it was ensured that the used words/phrases were removed from the search engine before conducting a second search round, so as not to affect the search results. That is, this procedure was meant for getting new articles specific for the search words/phrases of every search process.

### 4.2. Data refinement

Following data collection, a refinement process was carried out, excluding the irrelevant data. The excluded data contain:

- Enriched cited references = 2314
- Review articles = 696
- Open publisher-invited review =10
- Editorial materials = 38
- Letter = 32.

The excluded data also encompassed 4 meeting abstracts, 4 book reviews and 2 corrections. Note that the category *enriched cited references* contains the largest number of excluded articles, which can be ascribed to the fact that we used “All fields” in WoSCC search, which includes search terms found in titles, abstracts, author keywords, etc., thus resulting in a large number of enriched cited references. The irrelevant data were excluded right from the search strategy, i.e., these data were automatically excluded from WoSCC due to excluding them by WoS exclusion parameter (cf. Liu 2021). The refinement process also involved a manual exclusion. Put simply, after electronically screening the data, there remain also some articles that were misclassified as articles by WoS search engine, which were manually excluded. There were also articles miscategorized as articles, specifically those including the term “survey” but not “review” in their titles. This is perhaps the reason why WoS engine was not able to recognize them as reviews.

### 4.3 Data analysis

The study data were analyzed utilizing CiteSpace and VOSviewer programs. CiteSpace was used to generate clusters, calculate scientometric indicators including Modularity Q, Silhouette (S) values, and to analyze DCA, citation counts, burstness, betweenness centrality. CiteSpace parameters can be set to default settings (but they may be modified/changed, specifically in old versions of CiteSpace). The Q value ranges between 0 and 1, S value between -1 and 1, and g-index k between 25 and 100. Q and S are metric indicators, unveiling the quality and reliability of the clusters created, the clusters each represent a trending issue of research (Chen 2006). The best values of Q and S are the nearest to 1 indicating that the generated clusters are well-defined, and homogenous (Chen 2006, 2017; Ballouk et al. 2024).

In our study, given that CiteSpace 6.3.R1 is advanced software, the default settings were employed. As can be elicited from Fig. 2, the clustering process was conducted using the following default settings: g-index (k=25), LRF = 2.5, L/N=10, LBY=5, c=1.0, Q>0.8, S>0.8 as the selection criteria for identifying highly cited (and citing) references. This ensures that both highly and moderately cited (and citing) articles are included in the analysis, leading to a more comprehensive representation of AIL landscape, emerging issues and hotspot. Latent Semantic Indexing (LSI) algorithm was used as the primary clustering method, as it effectively captures conceptual similarities between documents. The use of Modularity and Silhouette validates the quality of the clusters. Higher Modularity values indicate well-isolated clusters (Chen 2003, 2006), while higher Silhouette scores reflect greater internal consistency within each cluster. Pruning was set to None, which helps reduce noise while preserving key structural relationships in the network (Chen 2006). The setting of the Look Back Year to 5-year intervals ensures that temporal trends in research evolution could be observed effectively. In presenting identified major/top clusters, we used the label Log-likelihood Ratio (LLR) algorithm-based keyword extraction (Tables 2 & 3), providing meaningful descriptions of thematic areas. The top clusters were further analyzed based on their representative articles and their role in shaping AIL research tending issues, and emerging hotspots, pinpointing AIL future frontiers.

In this study, there are 5750 publications in AIL research. The contributors of these publications are 2123 journals, 19664 authors, 5580 research centers, and 116 countries (extracted from VOSviewer). Fig. 1 depicts AIL publications along with g-index over the study timespan.

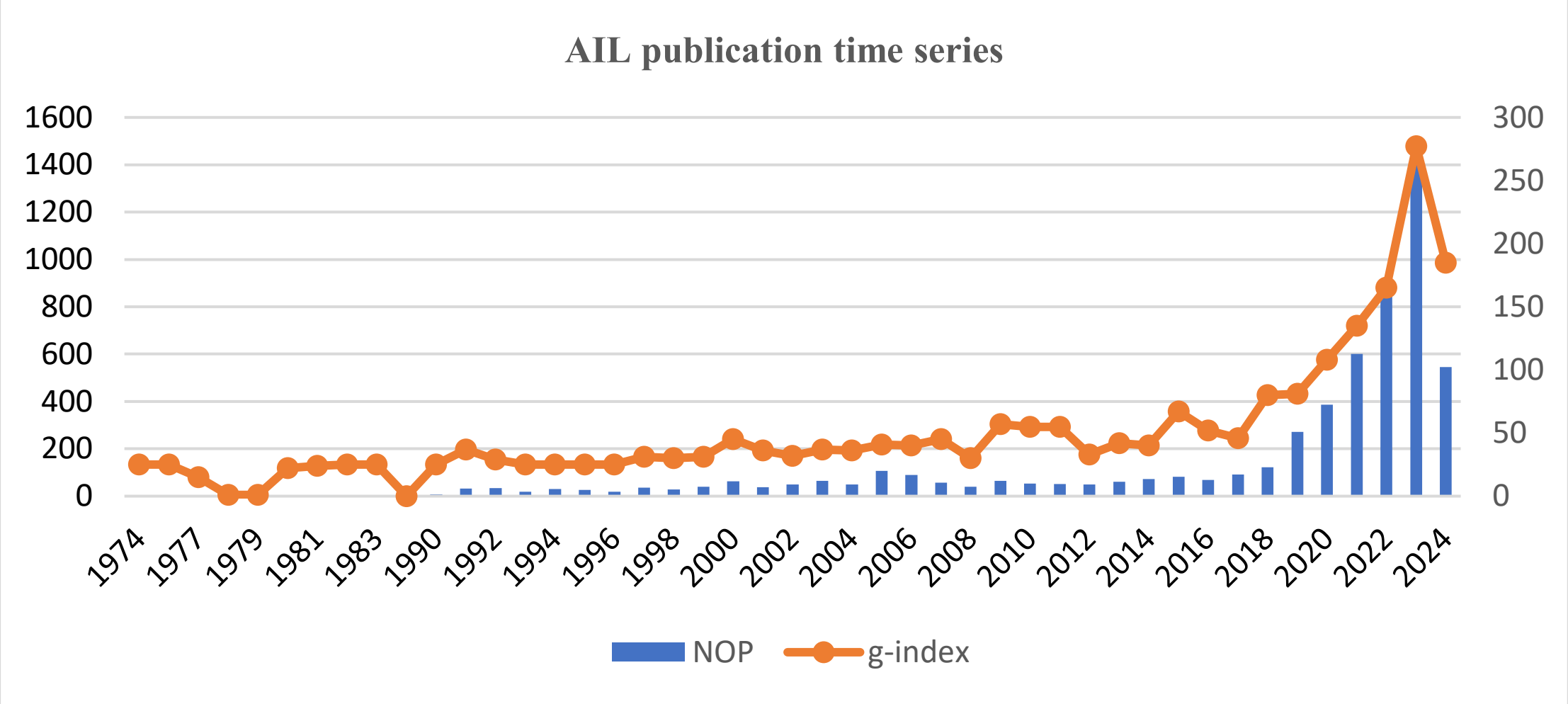

***Fig. 1: NOP and g-index increase in 1974-2024 timeframe***

As can be clearly seen in Fig. 1, AIL publication in its early years was considerably low. It was unstable, too: some time increases and some other time decreases. The same thing applies to g-index. g-index is a bibliometric measure designed to evaluate the scientific impact of a period of publication, author, journal. It is somehow different from, but an alternative citation measure to h-index. g-index can be set to 50 or 100, but as noted above, we used k = 25 as the default setting of CiteSpace. Except for 2005, the NOP was below 100 articles, upwards until 2017. We also notice that from 2018 onwards, i.e. in the 2018-2024 timeframe, AIL publication increase steadily. As Fig. 1 displays, AIL publication reaches its climax in 2023 with 1478 articles. Although the year 2024 is not complete at the time of data collection, AIL publication scores 546 articles only in January-March period, which signals AIL research considerable increase.

## 4. Results

### 4.1. 1974-2024 timeframe

In this timeframe, our aim was to provide a comprehensive picture of the intellectual knowledge of AIL research, pinpointing the trending issues and the (re)emerging hotspots. In this timeframe, the merged network consists of 427 clusters, 909 nodes, 5108 links, and 213437 citing and cited articles. The following are the values of these indicators for the whole network: (Q= 0.8958, S= 0.9433, Density= 0.0028, Harmonic mean {Q,S} = 0.9189). Table 2 displays the cluster information of the top 10 largest clusters.

**Table 2: Cluster information of top 10 largest cluster (1974-2024 timeframe)**

| ClusterID | Label (LLR) | Size | Silhouette | Average Year |
|---|---|---|---|---|
| 0 | *Using ChatGPT* | 205 | 0.894 | 2023 |
| 1 | *Interval-valued Intuitionistic Multiplicative Linguistic Preference Relation*[a] | 189 | 0.918 | 2017 |
| 2 | *Making Method* | 101 | 0.993 | 2016 |
| 3 | *Group Decision* | 78 | 0.982 | 2010 |
| 4 | *Explainable Artificial Intelligence* | 66 | 0.964 | 2017 |
| 9 | *Speaking Skill* | 45 | 0.976 | 2020 |
| 10 | *Computing Word Relatedness* | 40 | 0.998 | 2012 |
| 12 | *Visual Cluster* | 27 | 0.977 | 2019 |
| 25 | *Leveraging Tweet* | 11 | 0.996 | 2020 |
| 59 | *Ontology-Based Design Information Extraction* | 4 | 0.999 | 2003 |

[a]*Label has been overwritten using User-Defined Cluster Label Function in CiteSpace 6.3.R1*

In CiteSpace working mechanism, cluster labels represent trending issues (Chen 2003). Table 2 gives us a clear picture of the intellectual landscape of AIL research. In this timeframe, the first trending issue is *Using ChatGPT* (#0) with 205 members (Ms) and Silhouette (S) value of 0. 894, emerging around 2023 (i.e. the average year). The average emerging year is also illustrative of the launch of ChatGPT by OpenAI company in 2022 (Dergaa et al. 2023). This is very important; the trending issue- the intellectual knowledge produced- took almost one year after ChatGPT's launch. ChatGPT is perhaps the most recent development of AI deep learning models. The major citing article is Sohail et al. (2023). The second cluster (#1) is *Interval-valued intuitionistic multiplicative linguistic preference relation*. It has 189 Ms and 0.918 S, initiated around 2017. Rasmy et al. (2021) is the most citing article of the members of this cluster. *Making method* (#2) is the third trend of AIL research, having 101 Ms and 0.993 S. It emerged around 2016. The major citing article of the cluster is Wu et al. (2018). The fourth cluster (#3) is *Group Decision* with (78, Ms, 0.982 S). It evolved around 2010. The most citing article in this cluster is Zhang et al. (2018). The fifth (#4) cluster is *Explainable artificial intelligence* with (66 Ms, 0.964 S), emerging around 2017.

It is clear that the first five largest clusters each include more than 50 members, which is illustrative of these trends in AIL research in 1974-2024 timeframe. The next five clusters can be grouped as below 50 members each. These are *Speaking skill* (#9, 45 Ms and 0. 976 S), *Computing word relatedness* (#10, 40 Ms, and 0.998 S), *Visual Cluster* (#12, 27 Ms, and 0. 977 S), *Leveraging tweet* (#25, 11 Ms, and 0.996 S) and *Ontology-based design information extraction* (#59, 4 Ms, and 0.999 S).

These clusters are only the largest ones out of 427 clusters identified by CiteSpace in the 1974-2024 timeframe. It is important to note that the average years are above 2003, reflecting the publication development over years (Fig. 1). Fig. 2 depicts the Cluster view, and Fig. 3 the Timeline view of this timeframe.

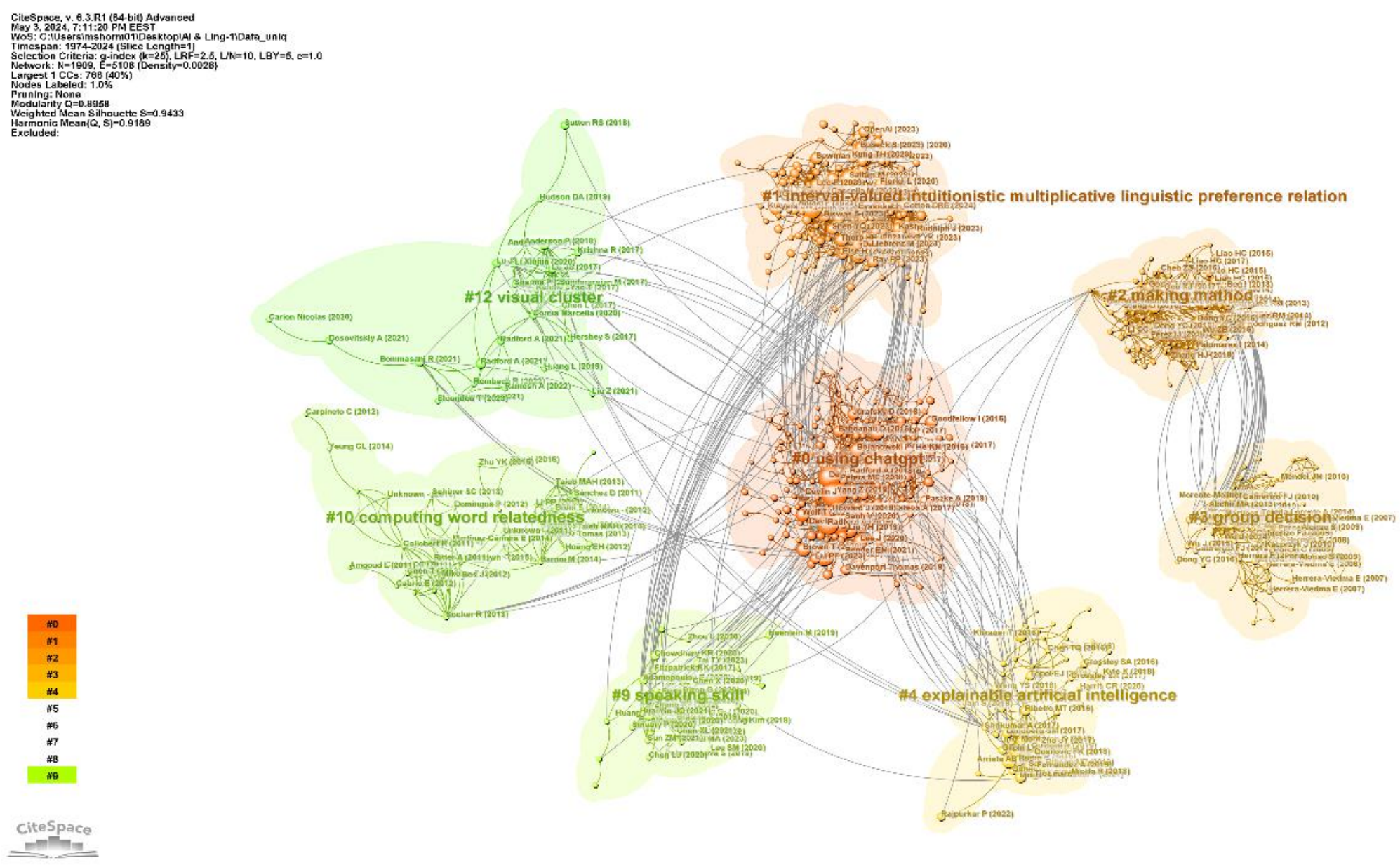

*Fig. 2: Cluster view of AIL trending issues in 1974-2024 timeframe*

Fig. 2 displays the Cluster view which provides critical insights into the structure and evolution of AIL knowledge production by identifying and visualizing clusters of related articles, thereby revealing thematic patterns within a domain (cf. Chen 2006). It helps detect emerging trends by analyzing how clusters evolve over time, offering a historical perspective on research development (Chen 2017). By visualizing research, it highlights connections between, for example, scholars and institutions, shedding light on influential networks in AIL research (cf. Chen and Song 2019). The visualization also uncovers key research articles that serve as pivotal nodes within a cluster, often representing seminal works that shape the trajectory of an academic discipline. For example, in Fig. 2 Cluster (#0), namely *Using ChatGPT* scores the highest number of publications, viz., 205. So, it is located at the center of the Cluster view (i.e. in the circle view).

Fig. 3 presents the Timeline view which provides a chronological perspective of AIL research development, allowing scholars to track the evolution of scientific knowledge over time by displaying clusters of related studies along a timeline (Chen 2006). In our study, the timeline view starts slightly earlier than our starting date, i.e. 1974 and ends in 2024. The thick circles indicate clusters or trending issues in AIL research with largest number of articles, which, in this case, is manifested by *Using ChatGPT.* This visualization method helps identify key turning points in a field by highlighting influential publications and their impact across different time periods, thus enabling users to analyze emerging trends, and hotspots in AIL knowledge production.

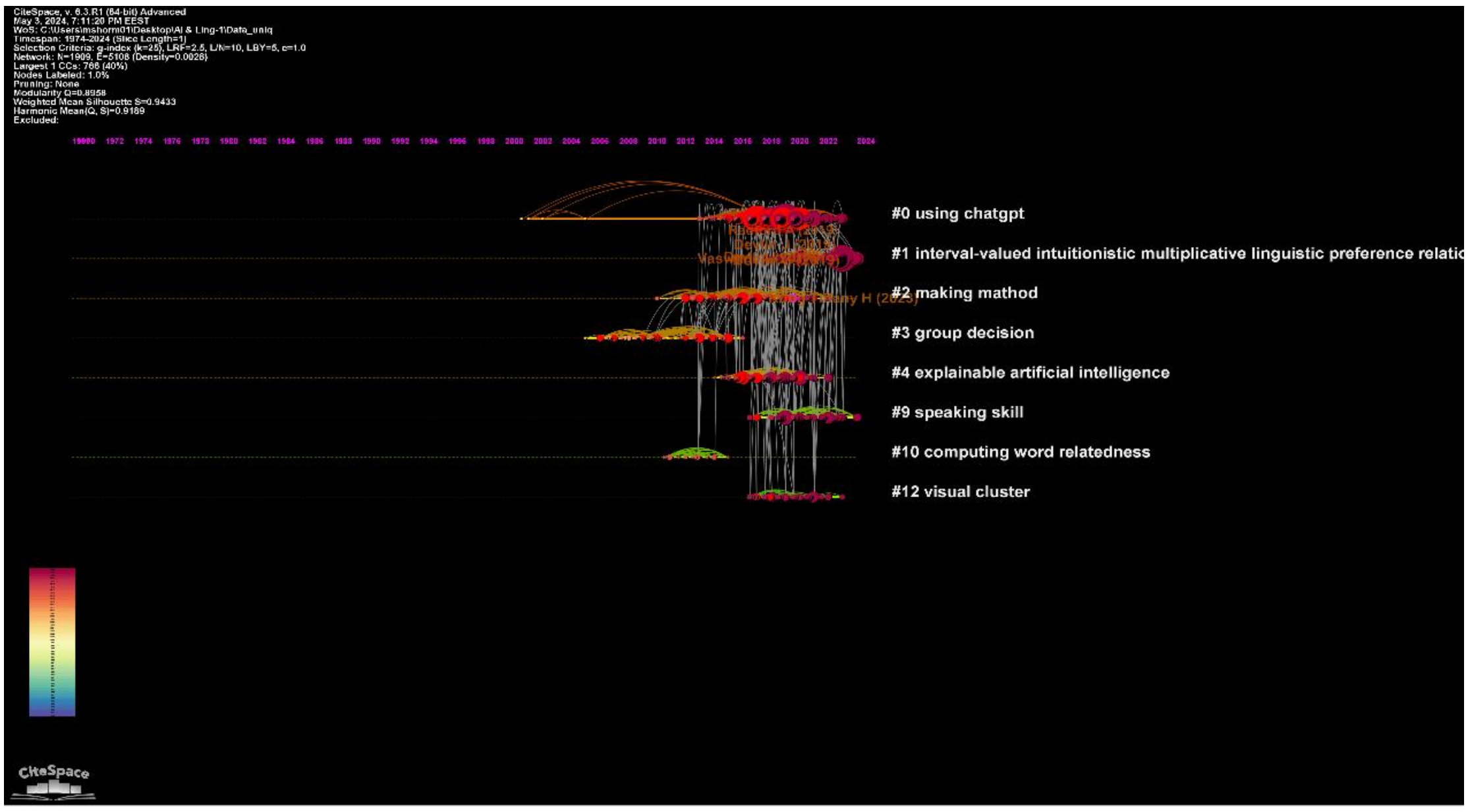

***Fig. 3: Timeline view of AIL trends in 1974-2024 timeframe***

Fig. 4 presents the top 12 cited articles in 1974-2024 timeframe, sorted by the strongest burst. The first reference with the strongest burst is Vaswani (2017) with burst value of 54.36 starting from 2021 and ending in 2022, followed by Devlin et al. (2019) with strongest burst of 35.91. Its burst spans between 2021-2022. The last cited article with strongest burst is Ribeiro (2016) with a burst value of 12.45, whose strength begins in 2020 and ends in 2021. CiteSpace provides a full picture for these articles in terms of burst, centrality, citation counts, degree, sigma.

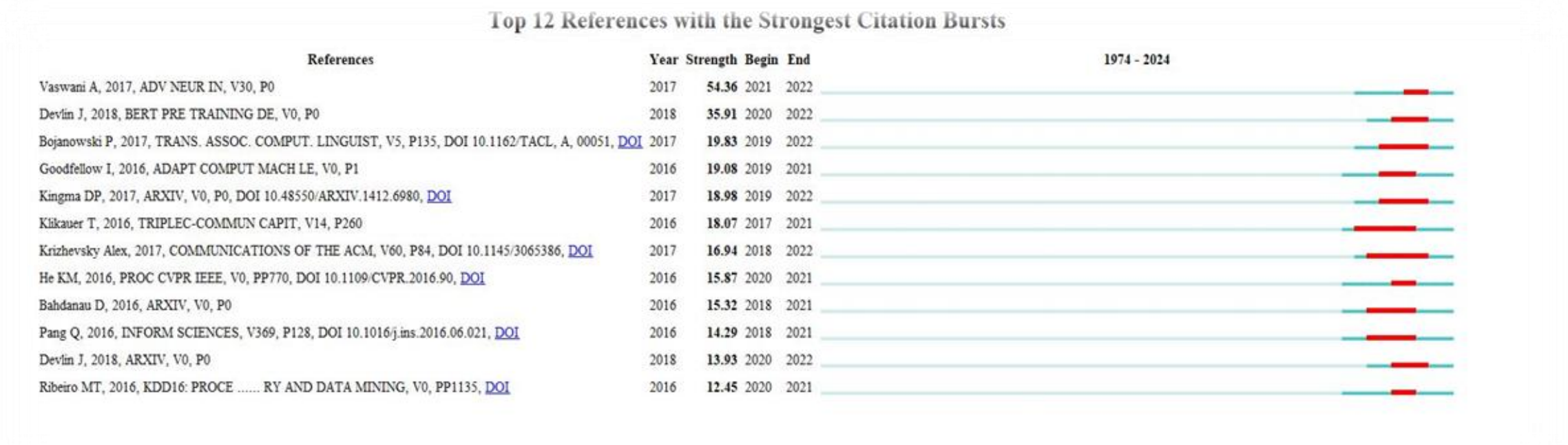

Top 12 References with the Strongest Citation Bursts

| References | Year | Strength | Begin | End | 1974 - 2024 |
|---|---|---|---|---|---|
| Vaswani A, 2017, ADV NEUR IN, V30, P0 | 2017 | **54.36** | 2021 | 2022 | |
| Devlin J, 2018, BERT PRE TRAINING DE, V0, P0 | 2018 | **35.91** | 2020 | 2022 | |
| Bojanowski P, 2017, TRANS. ASSOC. COMPUT. LINGUIST, V5, P135, DOI 10.1162/TACL, A, 00051, DOI | 2017 | **19.83** | 2019 | 2022 | |
| Goodfellow I, 2016, ADAPT COMPUT MACH LE, V0, P1 | 2016 | **19.08** | 2019 | 2021 | |
| Kingma DP, 2017, ARXIV, V0, P0, DOI 10.48550/ARXIV.1412.6980, DOI | 2017 | **18.98** | 2019 | 2022 | |
| Klikauer T, 2016, TRIPLEC-COMMUN CAPIT, V14, P260 | 2016 | **18.07** | 2017 | 2021 | |
| Krizhevsky Alex, 2017, COMMUNICATIONS OF THE ACM, V60, P84, DOI 10.1145/3065386, DOI | 2017 | **16.94** | 2018 | 2022 | |
| He KM, 2016, PROC CVPR IEEE, V0, PP770, DOI 10.1109/CVPR.2016.90, DOI | 2016 | **15.87** | 2020 | 2021 | |
| Bahdanau D, 2016, ARXIV, V0, P0 | 2016 | **15.32** | 2018 | 2021 | |
| Pang Q, 2016, INFORM SCIENCES, V369, P128, DOI 10.1016/j.ins.2016.06.021, DOI | 2016 | **14.29** | 2018 | 2021 | |
| Devlin J, 2018, ARXIV, V0, P0 | 2018 | **13.93** | 2020 | 2022 | |
| Ribeiro MT, 2016, KDD16: PROCE ...... RY AND DATA MINING, V0, PP1135, DOI | 2016 | **12.45** | 2020 | 2021 | |

***Fig. 4: Top 12 cited articles by bursts in 1974-2024 timeframe***

## 4.2. 2018-2024 timeframe

In 2018-2024 timeframe, we wanted to characterize the research trends, hotspots in the last 7 years. We intend in this timeframe to examine the increase of publication, where it increases noticeably crossing hundred publications annually (Fig. 1), so that the intellectual knowledge production in this period is effectively measured, evaluated and analyzed, uncovering its strengths and weaknesses. To begin with, the merged network in 2018-2024 timeframe has 83 clusters, 848 nodes, 2820 links, and 184306 citing and cited references. In this timeframe, the values of the scientometric indicators for the whole network are as follows: (Q= 0.7399, S= 0.9051, Density= 0.0095, Harmonic mean {Q,S} = 0.8142).

**Table 3: Cluster information of top 11 largest clusters (2018-2024 timeframe)**

| ClusterID | Label (LLR) | Size | Silhouette | Average Year |
|---|---|---|---|---|
| 0 | Natural Language Processing | 110 | 0.883 | 2016 |
| 1 | *Cross-sectional Study* | 103 | 0.887 | 2022 |
| 2 | *Making Method* | 98 | 0.962 | 2016 |
| 3 | *Academic Writing* | 85 | 0.864 | 2022 |
| 4 | *Using Bidirectional Encoder Representation* | 78 | 0.842 | 2019 |
| 5 | *Explainable Artificial Intelligence* | 50 | 0.939 | 2018 |
| 6 | *Speaking Skill* | 44 | 0.961 | 2020 |
| 7 | *Novice Programmer* | 34 | 0.917 | 2019 |
| 8 | *Artificial Intelligence* | 26 | 0.962 | 2016 |
| 9 | *Prioritization* | 7 | 1 | 2014 |
| 10 | *Context-Based Fake News Detection Model* | 5 | 0.998 | 2019 |

Compared to Table 2, Table 3 clearly illustrates the change of emerging issues, hotspots, and more importantly, the reemergent/recurrent issues and hotspots in AIL research. The first trending issue in this timeframe is *Natural Language Processing* (#0) with 110 Ms, 0.883 S, emerging around 2016. We notice that including (#0), there are 8 (newly) emergent trending issues in AIL research in 2018-2024 timeframe: *Cross-sectional Study* (#1, 103 Ms, 0.887 S), *Academic writing* (#3, 85 Ms, 0.864 S), *Using bidirectional encoder representation* (#4, 78 Ms, 0.842 S), *Novice programmer* (#7, 34 Ms, 0.917 S), *Artificial intelligence* (#8, 26 Ms, 0.962 S), *Prioritization* (#9, 7 Ms, 1 S), and *Context-based fake news detection model* (#10, 5 Ms, 0.998 S). Note that the trending issue *Using bidirectional encoder representation* (from Transformer) BERT, we have discussed so far. These newly emerging clusters each represent newly emerging research trends of AIL in 2018-2024 timeframe.

However, there are 3 remerging trending issues which are represented by *Making method* (#2, 98 Ms, 0.962 S), *Explainable artificial intelligence* (#5, 50 Ms, 0.939 S), and *Speaking skill* (#6, 44 Ms, 0.961 S). Except *Making method* (#2), these remerging AIL trending issues each have undergone a change of the status/rank, size of members, S, and average year in this timeframe. For example, *Explainable artificial intelligence* was the fifth cluster (#4) in 1974-2024 timeframe, but it is the $6^{th}$ (#5) in 2018-2024 timeframe. The major citing articles of this timeframe include Porcel et al. (2018), Howard (2019), Zhang et al. (2018), Sohail et al. (2023), Kolides et al. (2023), Li et al. (2023), Jeon and Lee (2023), and Brinkmann et al. (2023).

Notice that in 1974-2024 timeframe, *Using ChatGPT* is the largest cluster, a purely AI trending issue, while in 2018-2024 timeframe, the largest cluster is *Natural language processing*, a purely (computational) linguistics trending issue. This in a way or another reflects the correlation between linguistics and AI.

Figs. 5 & 6 depict the Cluster view and Timeline view of 2018-2024 timeframe, respectively.

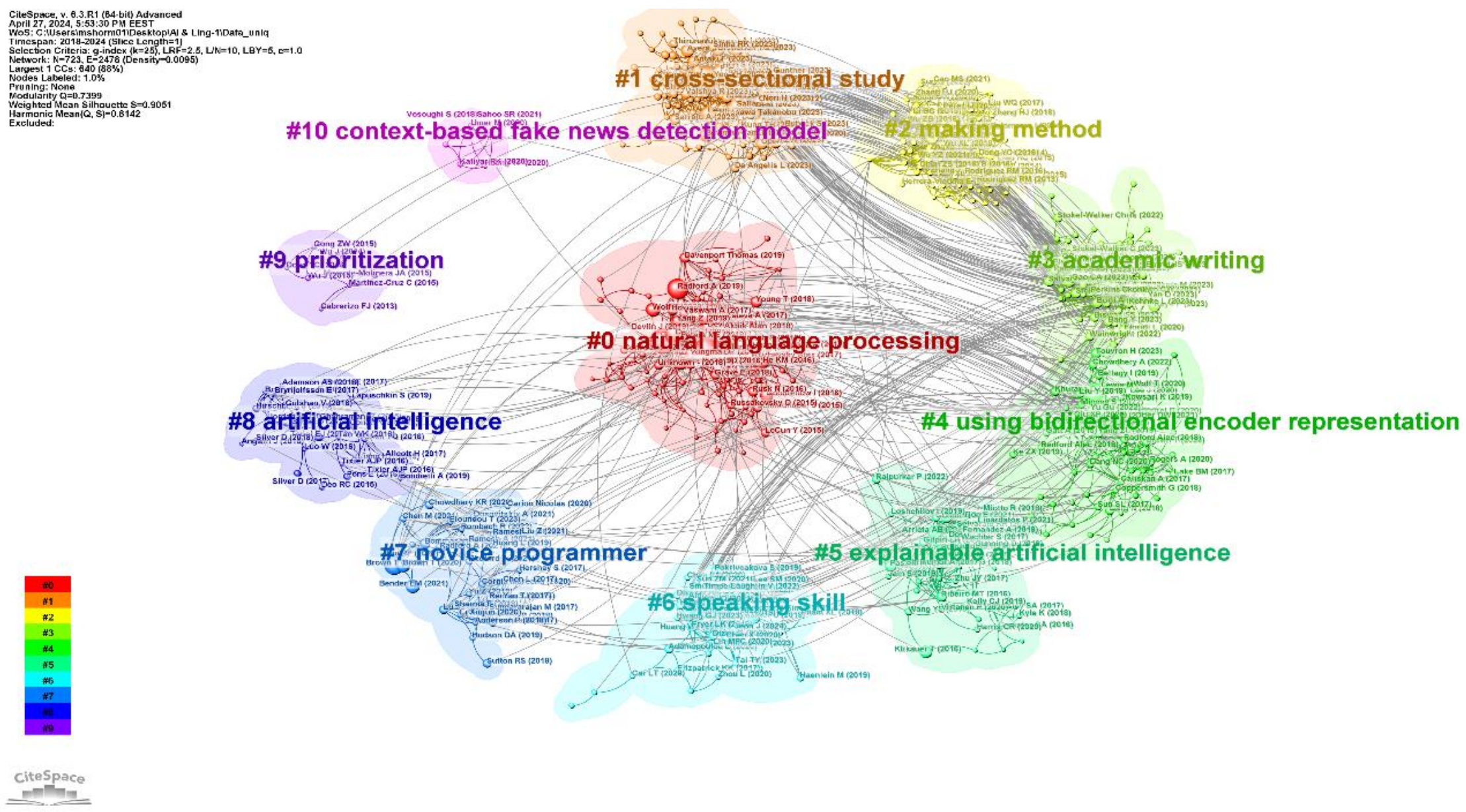

*Fig. 5: Cluster view of AIL trends in 2018-2024 timeframe*

In Fig. 5, we find that those clusters with the largest number of articles (cf. Table 3) occupy the center of the circle view. For example, (#0) *Natural Language Processing* has 110 articles. It has the highest publication impact in this timeframe, in contrast to *Using ChatGPT* in 1974-2024 timeframe. Due to the short period of the timeframe, i.e. 7 years from 2018 to 2024, limited number of articles, small numbers of clusters (83) nodes (848), links (2820), and citing and cited references (184306), it seems that the Cluster view in Fig. 5 is clearer than that of 1974-2024 timeframe (cf. Fig. 2). The same thing can also be noticed in Fig. 6 below, which displays the Timeline view of AIL research in 6-year period.

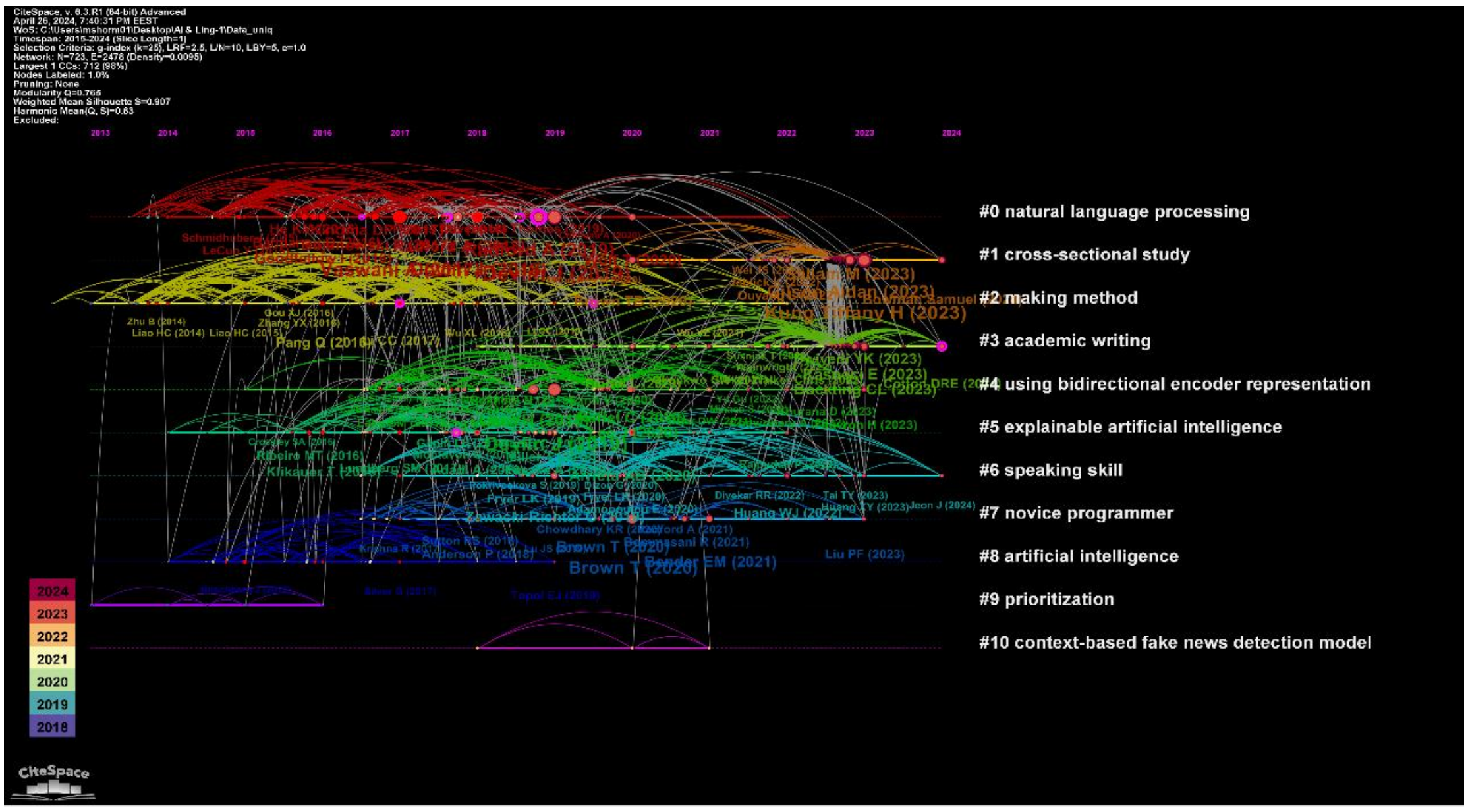

*Fig. 6: Timeline view of AIL trends in 2018-2024 timeframe*

In this Timeline view, mapping visualizations of AIL trending issues and hotspots get clearer, harmonized and well-defined (Figs. 5 & 6). This is very clear from comparing Figs. 2 & 3 to Figs. 5 & 6.

Fig. 7 presents the DCA of the top 12 cited references in 2018-2024 timeframe, sorted by the strongest burst. The first three references with the strongest burst are Vaswani (2017), Dovlin et al. (2018), and Goodfellow et al. (2016). Their burst values are 40, 28.42, and 17.8, respectively. However, the duration of their burst differ as can be seen in Fig. 7.

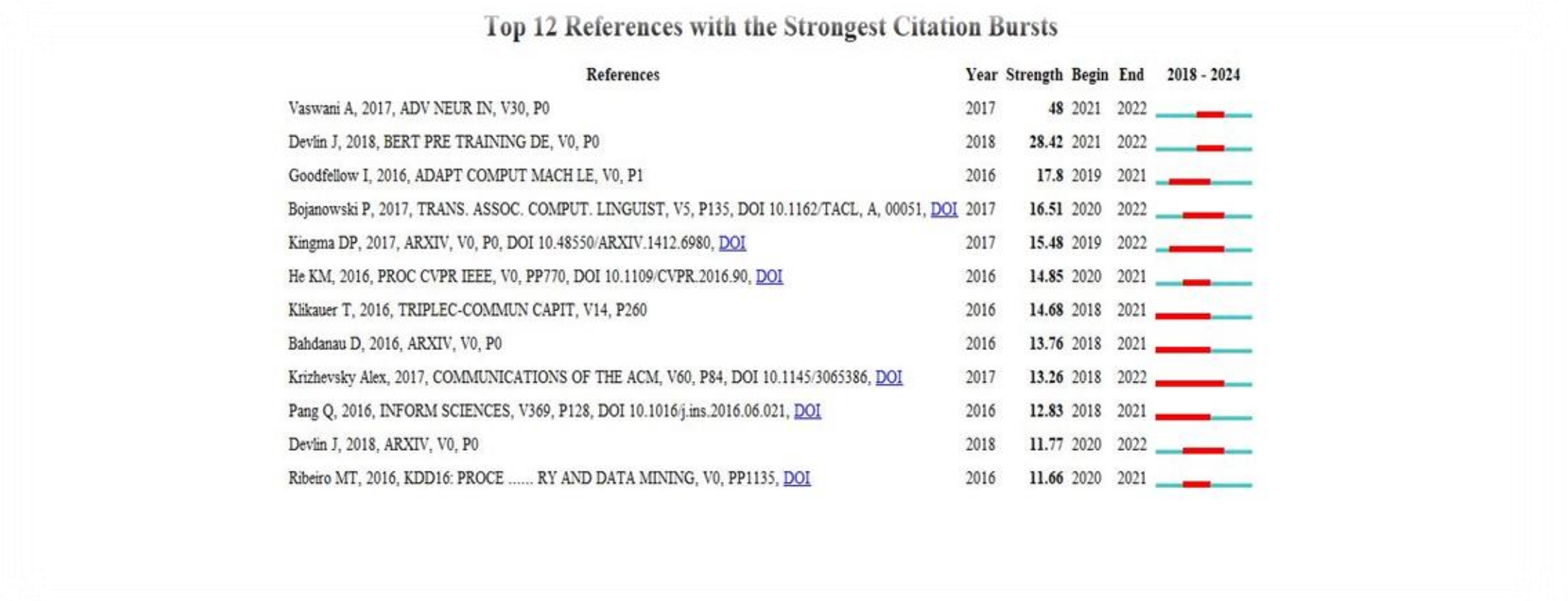
Top 12 References with the Strongest Citation Bursts

| References | Year | Strength | Begin | End | 2018 - 2024 |
|---|---|---|---|---|---|
| Vaswani A, 2017, ADV NEUR IN, V30, P0 | 2017 | 48 | 2021 | 2022 | |
| Devlin J, 2018, BERT PRE TRAINING DE, V0, P0 | 2018 | 28.42 | 2021 | 2022 | |
| Goodfellow I, 2016, ADAPT COMPUT MACH LE, V0, P1 | 2016 | 17.8 | 2019 | 2021 | |
| Bojanowski P, 2017, TRANS. ASSOC. COMPUT. LINGUIST, V5, P135, DOI 10.1162/TACL, A, 00051, DOI | 2017 | 16.51 | 2020 | 2022 | |
| Kingma DP, 2017, ARXIV, V0, P0, DOI 10.48550/ARXIV.1412.6980, DOI | 2017 | 15.48 | 2019 | 2022 | |
| He KM, 2016, PROC CVPR IEEE, V0, PP770, DOI 10.1109/CVPR.2016.90, DOI | 2016 | 14.85 | 2020 | 2021 | |
| Klikauer T, 2016, TRIPLEC-COMMUN CAPIT, V14, P260 | 2016 | 14.68 | 2018 | 2021 | |
| Bahdanau D, 2016, ARXIV, V0, P0 | 2016 | 13.76 | 2018 | 2021 | |
| Krizhevsky Alex, 2017, COMMUNICATIONS OF THE ACM, V60, P84, DOI 10.1145/3065386, DOI | 2017 | 13.26 | 2018 | 2022 | |
| Pang Q, 2016, INFORM SCIENCES, V369, P128, DOI 10.1016/j.ins.2016.06.021, DOI | 2016 | 12.83 | 2018 | 2021 | |
| Devlin J, 2018, ARXIV, V0, P0 | 2018 | 11.77 | 2020 | 2022 | |
| Ribeiro MT, 2016, KDD16: PROCE ...... RY AND DATA MINING, V0, PP1135, DOI | 2016 | 11.66 | 2020 | 2021 | |

***Fig. 7: Top 12 cited articles by burst in 2018-2024 timeframe***

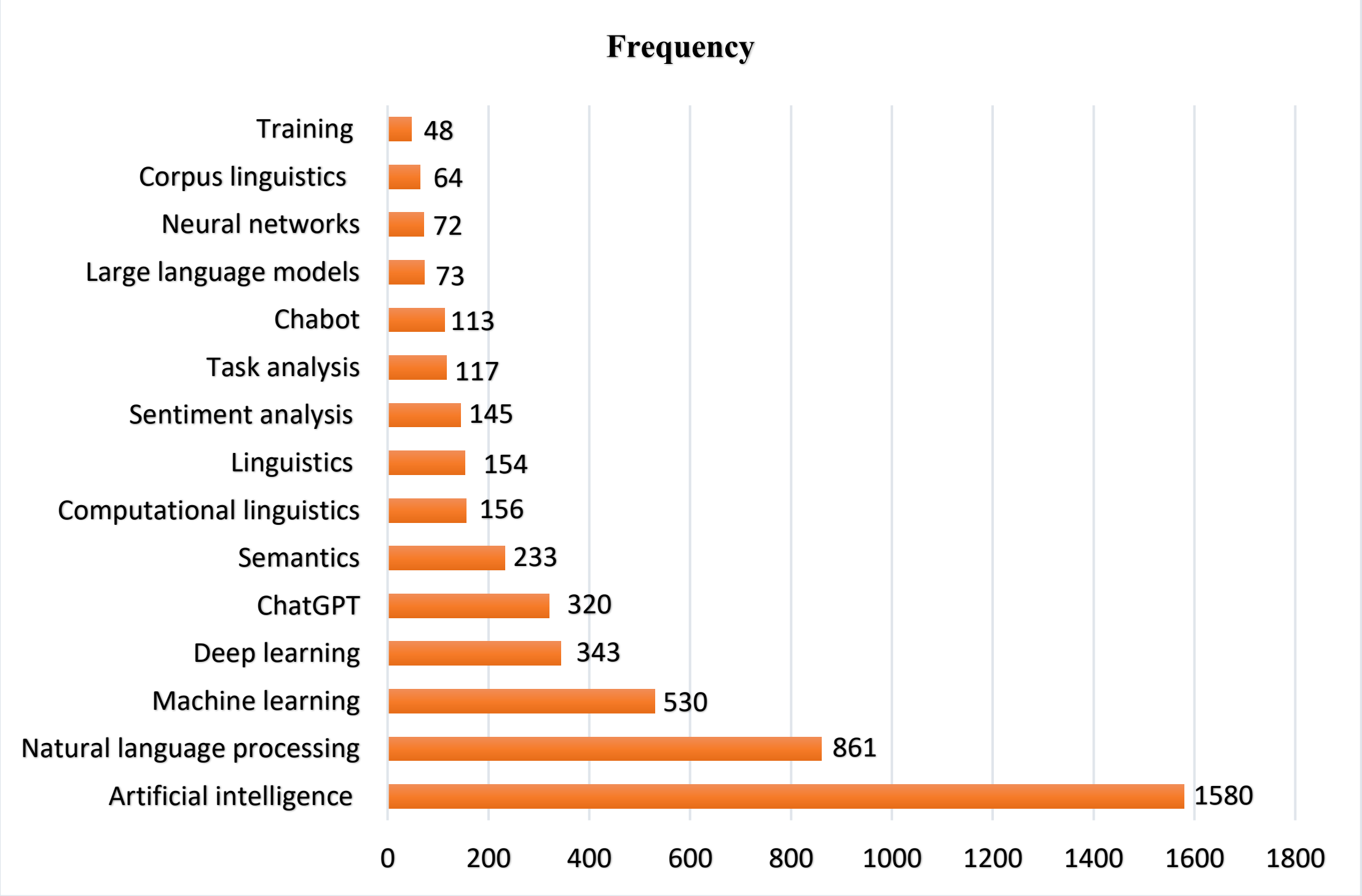

***Fig. 8: Top 15 author's keywords***

In this study, there are 15046 author's recurrent keywords or KCA, but only 665 meet the VOSviewer's threshold. Fig. 8 showcases the top 15 keywords. These keywords are by and large consistent with our cluster/trend analysis. The keyword *Artificial intelligence*, with 1580 Freq retains the first rank. *Natural language processing* with 861 Freq occupies the second rank. The third and fourth ranks are retained by *Machine* *learning* and *Deep learning* with 530 and 343 frequencies, respectively. ChatGPT occupies the fifth rank with 320 Freq.

There are 6 keywords related to linguistics, namely: *Natural language processing* (861 Freq, second rank)*, Semantics* (233 Freq, sixth rank)*, Computational linguistics* (156 Freq, seventh rank)*, Linguistics* (154 Freq, eighth rank), *Sentiment analysis* (145 Freq, ninth rank), and *Corpus linguistics* (64 Freq, fourteenth rank).

The rest top recurrent keywords belong to AI including *Artificial intelligence* (1580 Freq, first rank), *Machine learning* (530 Freq, third rank)*, Deep learning* (343 Freq, fourth rank)*, ChatGPT* (320 Freq, fifth rank), *Large language models* (114 Freq, eleventh rank) and *Chabot* (113 Freq, twelfth rank), *Corpus linguistics* (64 Freq, fourteenth rank). The fifteenth keyword *training* with 48 Freq may be connected to DNNs, ANNs or training ChatGPT on massive data, or it may be connected to training on *Speaking skill* as in the cluster (#6).

Fig. 9 portrays mapping of author's keywords. The information displayed unveils a very important finding which, in a way or another, signals the correlation between linguistics and AI in AIL research.

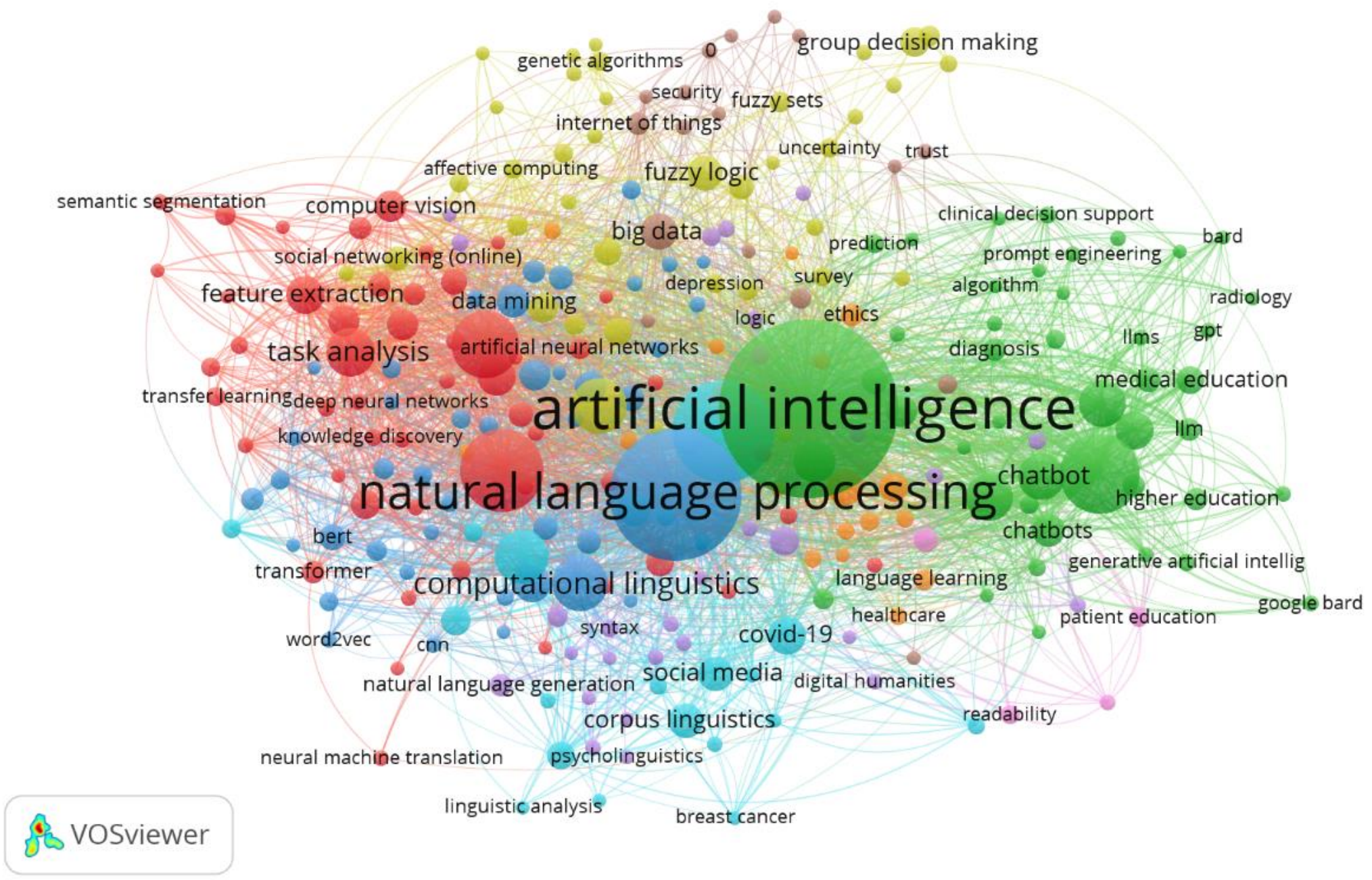

***Fig. 9: Mapping of author's keywords***

In Fig. 9, generated by VOSviewer, mirrors Fig. 8, in that, *Artificial intelligence* seems to be the "biggest" node representing the top most co-recurrent author's keywords in AIL research. It is followed by *Natural language processing* and so on. In fact, Fig. 9 displays only 665 (out of 15046) keywords. Overall, the keywords presented in Figs. 8 and 9 give us a vivid clue that AIL authors pay much attention to these keywords while researching linguistics and AI correlation. Additionally, Fig. 9 tells us that our search strategy also ends up with keywords such as *Google bard, Neural machine translation* which may be connected to Machine Translation (MT) industry, and the developments it has undergone (see also Shormani 2024d). Keywords like *Linguistic analysis, Corpus linguistics, Computation linguistics, Large language learning, etc.* reflect the contribution of linguistics to AI. A substantial trending issue of AIL research that could be highlighted here is the fact that AIL research not only focuses on the correlation between linguistics and AI, but it also shows that AIL contributes to many other fields including medicine as evident by the keyword *Breast cancer.*

### 4.3. Key contributors to AIL research

In this section, we tackle key contributors to AIL knowledge production by addressing journals, research centers/universities, and countries.

#### 4.3.1. Top key productive journals

Recall that in our study there are 2123 journals, but only 223 meet VOSviewer's thresholds. Fig. 10 depicts the top 12 productive journals in AIL research in 1974-2024 timeframe, sorted by NOPs and DCA. *IEEE Access* is the most productive journal contributing to AIL research with 175 publications and 938 citations, followed by *Engineering Application of Artificial Intelligence* with 117 publications and 2178 citations. The third leading journal is *Artificial Intelligence Review* (69 publications, 1204 citations).

The last three top journals are *Fuzzy Sets and Systems* (245 publications, 2858 citations)*, International Journal of Intelligent Systems* (24 publications, 1243 citations)*,* and *Information Fusion* (22 publications, 2264 citations).

The three top ranks were occupied by *IEEE Access, Engineering Application of Artificial Intelligence* and *Artificial Intelligence Review,* demarcating the global status of these journals as the most leading contributing journals in AIL world. Almost all the journals reflect the areas of AIL research, which is consistent with the cluster analyses so far.

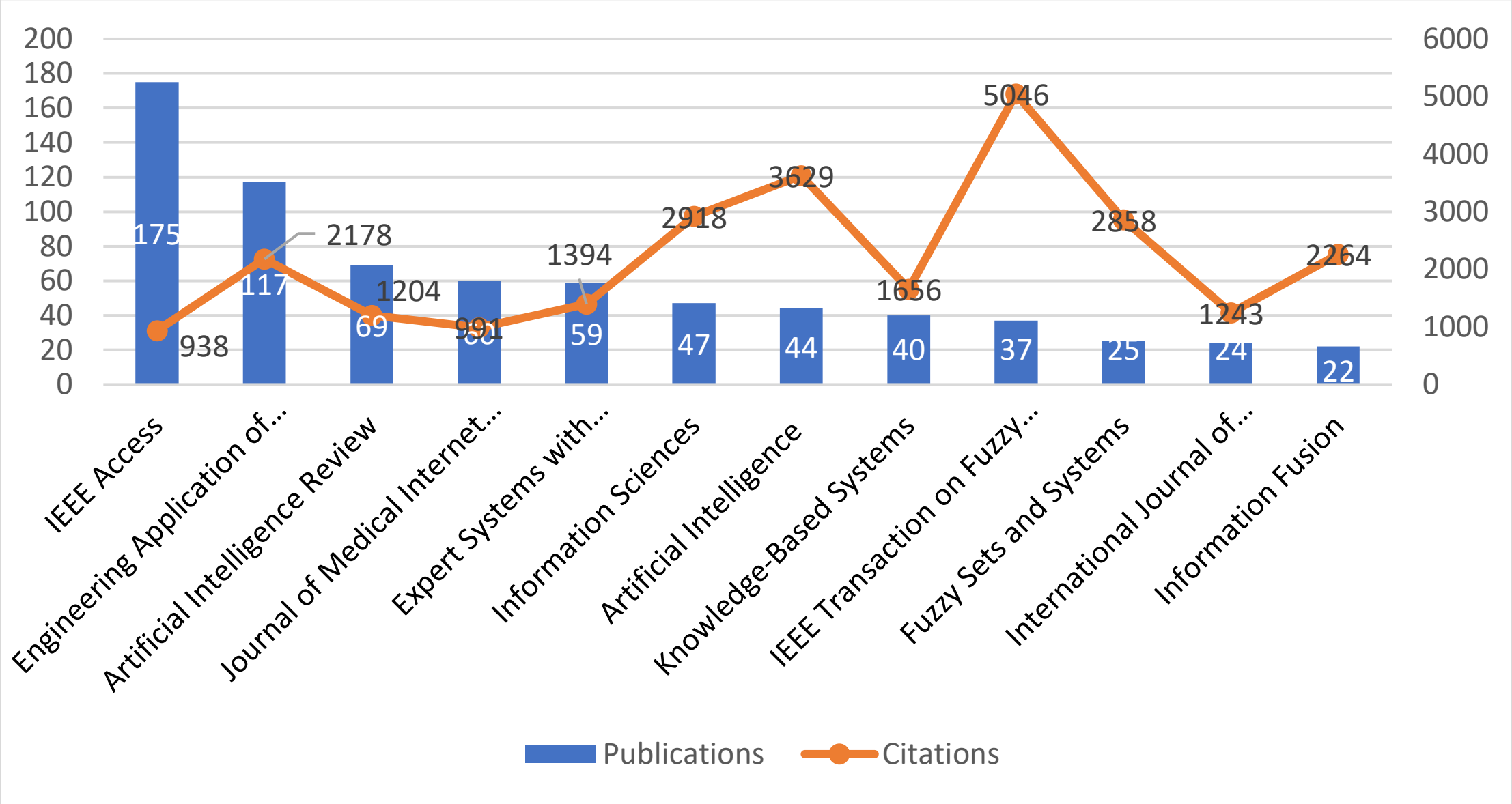

***Fig. 10: Key journals in AIL research***

Fig. 11 presents the knowledge mapping of key journal contributors to AIL intellectual production.

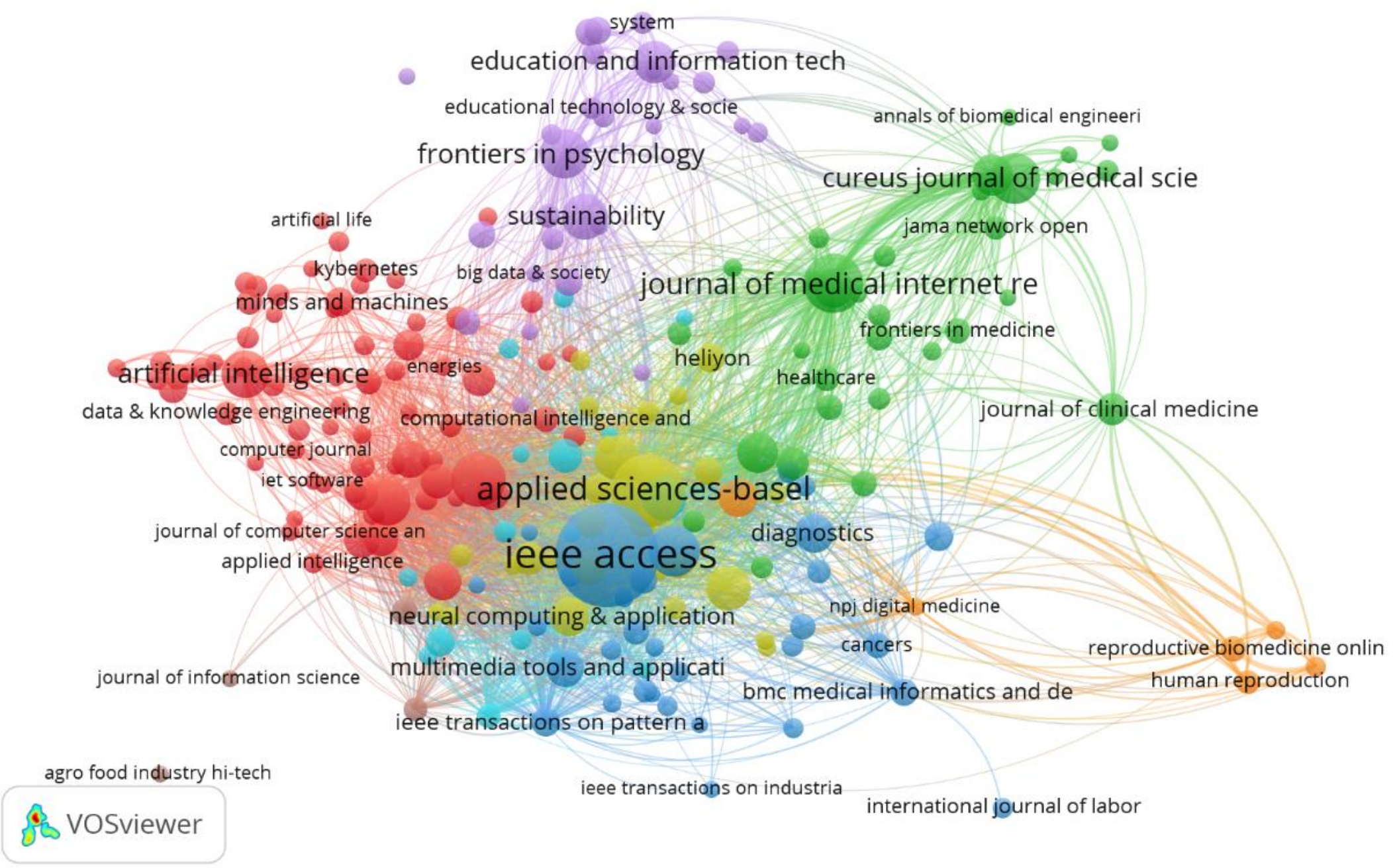

*Fig. 11: Mapping of key journals in AIL research*

Not only does Fig. 11 reflect the consistency of our analysis, it also shows high Factor Impact, largest downloads, largest NOPs, and highest DCA, thus guiding AIL researchers where to publish their studies. For example, the three leading journals that have these characteristics are *IEEE Access, Engineering Application of Artificial Intelligence,* and *Artificial Intelligence Review*. The journals in Figs. 10 and 11 may also continue this leading role for at least the 10 coming years.

### 4.3.2. Top key productive research centers

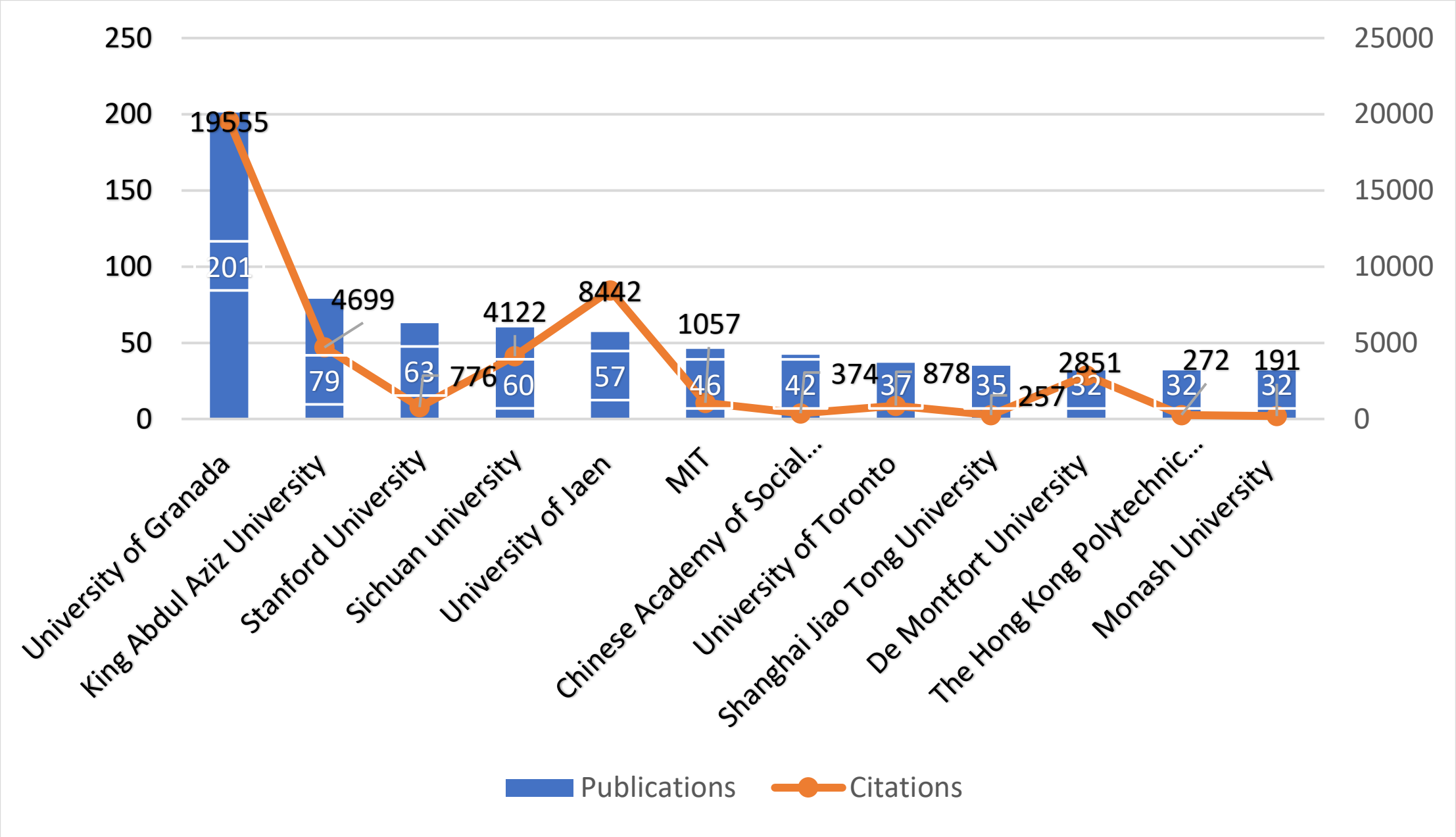

*Fig. 12: Key research centers in AIL*

There are 5580 institutions and research centers in our analysis, but only 554 meet VOSviewer's thresholds. The data showcased in Fig. 12 present the top 12 institutions and research centers sorted by NOP and DCA contributing to AIL research in the world. Scrutinizing Fig. 12, we are likely to find a panoramic picture of contributing research institutions. The top three research centers are University of Granada, with 201 publications and 19555 citations, enjoying the first rank. It is a Spanish University located in Granada. The second rank is retained by an Arab institution, namely King Abdul Aziz University. It has 79 publications and 4699 citations, as one of the top research centers in the Arab world. The third rank is occupied by Stanford University, which is a USA research center.

To summarize, among the top key institutional contributors, there are 4 Chinese universities, 3 USA universities, 2 Spanish universities, 1 Canadian university, namely University of Toronto and 1 UK research institution, namely De Montfort University, and 1 Arab University.

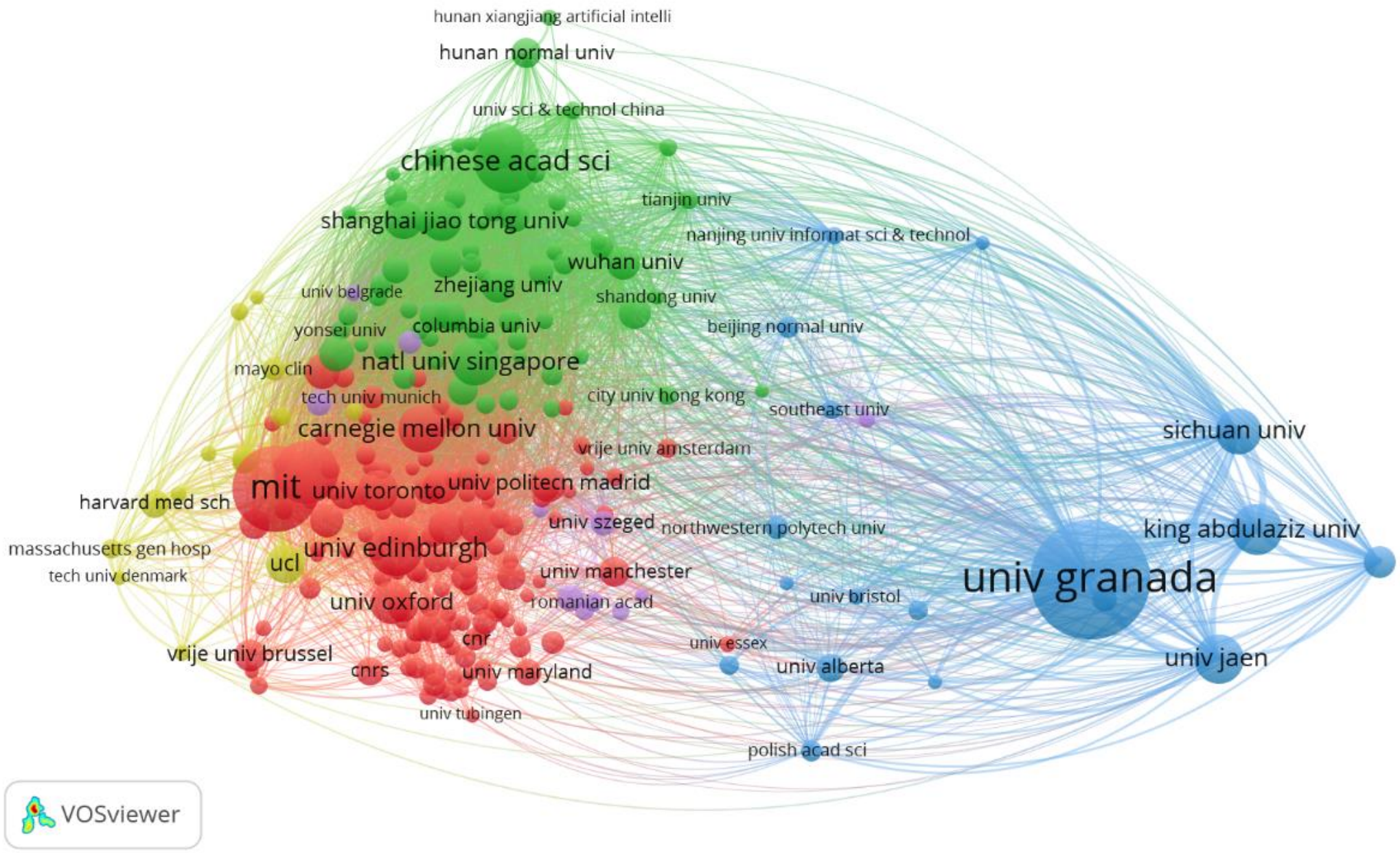

***Fig. 13: Mapping of research centers***

Fig. 13 gives us a full picture of research centers and institutions that contribute to AIL research current state. It shows also the possible leading research centers in the future which may continue this leading role. For example, University of Granada, King Abdul Aziz university, and Stanford University having the largest NOPs and the highest DCA may continue this leading role for at least the 10 coming years.

### 4.3.3 Top key productive countries

Recall that there are 116 countries contributing to AIL in our corpus. In terms of country contributors, the top 12 key leading countries contributing to AIL research seem to be consistent with the analysis of research centers and universities. The data displayed in Fig. 14 depict the top contributing countries to AIL research in the world. The first rank is retained by USA with 1177 publications and 21343 citations. The second rank is retained by China with 1070 publications and 15753 citations. Spain (519 publications and 23070 citations) ranks third. All in all, there are 6 Western countries, namely Spain,

England, Germany, Italy, Canada and France. There are 4 Asian countries namely China, India, Saudi Arabia, and South Korea. The two remaining countries are Australia and USA.

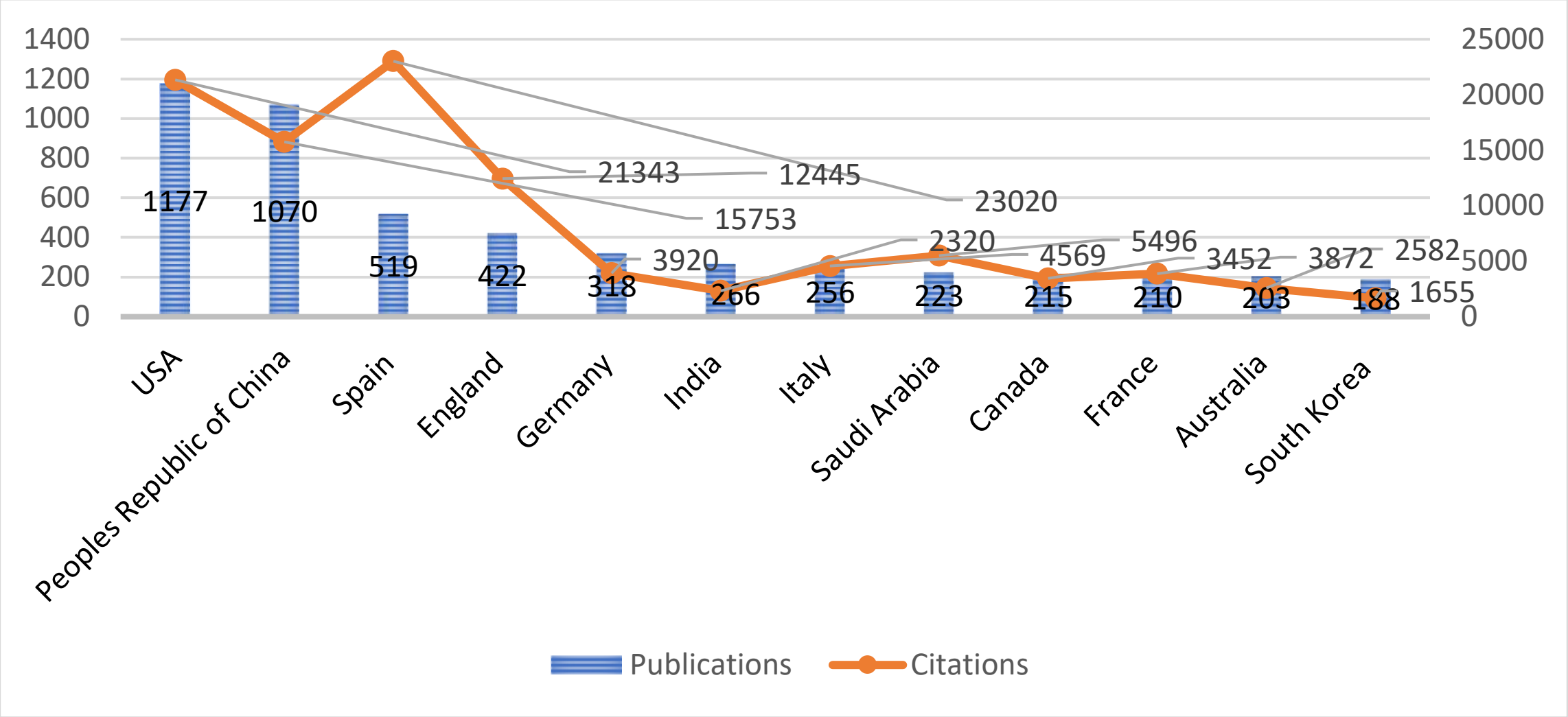

*Fig. 14: Key country contributors*

Fig. 15 below further supports our argument above. Apart from the top 12 countries contributing to AIL research, it reveals other countries, though with less contribution. Among the Asian countries, Bangladesh, Indonesia, and Pakistan play a considerable role in contributing to AIL research. European countries also include Cyprus, Switzerland, and Sweden. Finally, of the Arab countries, United Arab Emirates, Egypt, and Oman seem to play a significant role in contributing to AIL research. (Note that in Fig., VOSviewer labels both England and Scotland as country names, though they are within the United Kingdom).

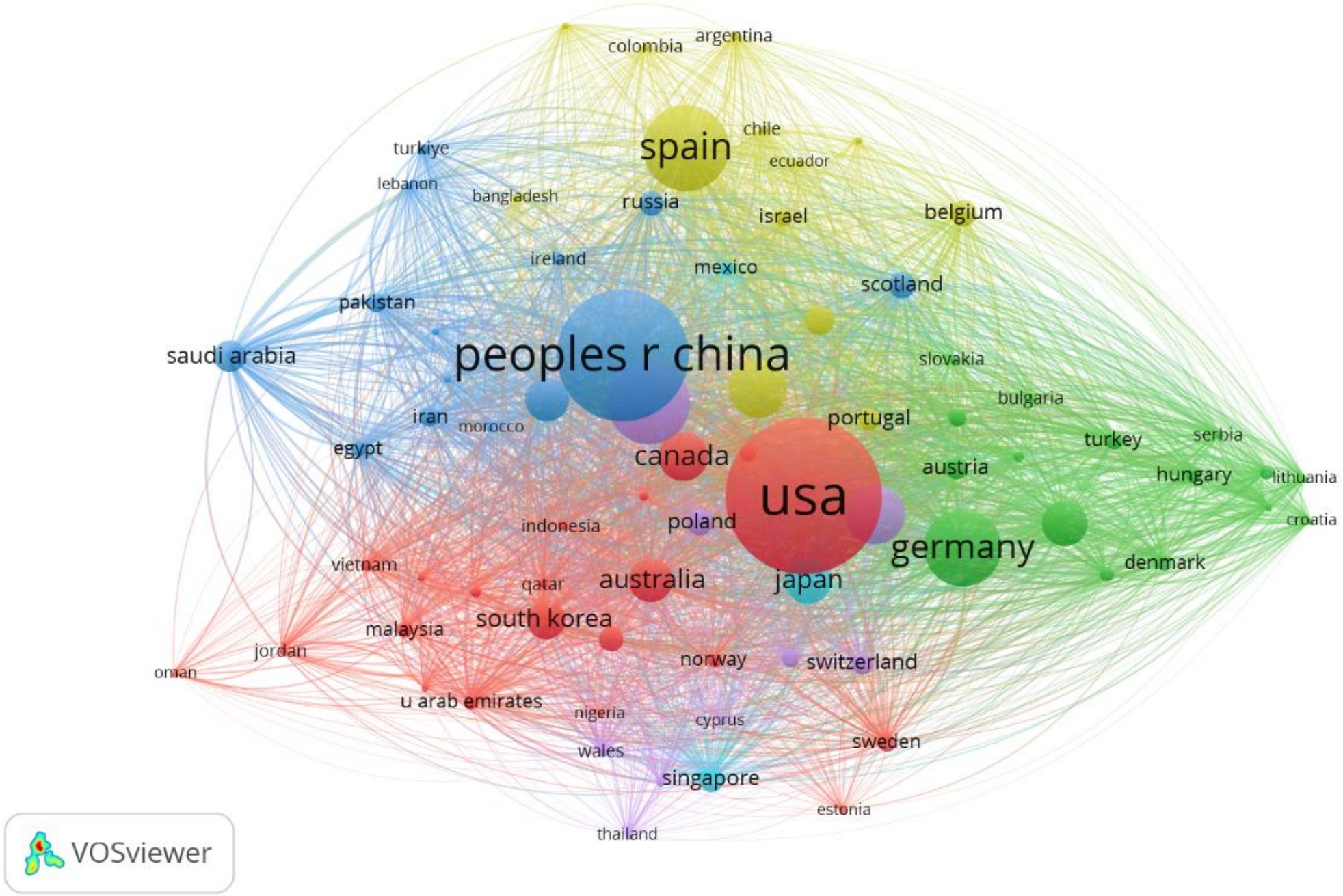

*Fig. 15: Mapping of key country contributors*

## 5. Discussion

The link between linguistics and AI spans about a century now. It takes the shape of providing substantial bases to AI by linguistics. The result of this interaction is giving birth to computational linguistics, a field of study that could be viewed as the most contributing body to AI. With this in mind, this study traces this link or interaction in more than half a century. In this study, there are 5750 WoSCC articles published in 2123 journals, which are written by 19664 authors belonging to 5580 research centers in 116 countries. In this very aspect, this study encapsulates the intellectual production in AIL research, unveiling the actual knowledge landscape and pinpointing the trending frontiers and emerging research hotspots in this area of study. These landscape, trends, and hotspots are clearly mapped and visualized. To begin with, Table 2 presents the publication development over time showing the g-index of publication by year (Fig.1). This gives us a steady background of how AIL research develops, uncovering the fact that in its early stages, AIL research was not strong. That only one article was published in 1974 clearly reflects this fact. However, the more we proceed in discovering Fig. 2, the publication increases until reaching its climax in 2023 with 1478 publications.

In 1974-2024 timeframe, there have appeared several and varied research trends. CiteSpace identities 427 research trends, and 213437 citing and cited articles. The most important of these research trends are 10 (Table 2) including *Using ChatGPT*, *Interval-valued intuitionistic multiplicative linguistic preference relation*, *Making method*, *Explainable artificial intelligence, Speaking skill*, *Computing word relatedness*, *Visual cluster, Leveraging Tweet,* and *Ontology-based design information extraction. Using ChatGPT* is the topmost research trend, emerging around 2023 and dominating AIL research scene, with 205 research articles as members of this research trend. ChatGPT (=Chat-Generative Pre-Trained Transformer) is perhaps the latest development of AI large language models, a deep learning model the main purpose of which was translation (Jiao et al. 2023; Siu 2023; van Dis et al. 2023; Shormani 2024c/d). In addition to translation, ChatGPT is a chatbot, designed to chat with humans, engaging in several types of conversation and for different purposes, leveraging "the power of GPT to provide interactive and dynamic responses, mimicking human-like conversation" (Sohail et al. 2023: 1, see also Shormani 2024c). It has been reported to perform competitive tasks, and sometimes surpasses human (Kung et al. 2023). Its first version is GPT-1, having 117 million parameters, and has been trained on massive amounts of data (Ernst and Bavota 2022; Sohail et al. 2023). For example, Sohail et al. (2023), the major citing article, uncovers how ChatGPT is encoded, providing a detailed map of existing research, current challenges to ChatGPT working realm and how and where future trends should be directed.

Note that in both timeframes, i.e. 1974–2024 and 2018–2024, ChatGPT seems to overlap the two timeframes, and the impact of ChatGPT influences both datasets. Given the historical contextualization, it is important to shed light on the scenario prior to ChatGPT, thereby clearly highlighting its impact. The pre-ChatGPT landscape (prior to late 2022), there are key developments in AI and NLP from 2018 to 2022, such as the rise of transformer-based models (e.g., BERT, GPT-1), as well as the broader discourse around machine learning in language generation and its early limitations. This contrasts with the post-ChatGPT period, i.e. post-2022. In this period, ChatGPT, and AI industry, in general, has witnessed vast and fast development, which is marked by an exponential surge in interest, publication volume, and cross-disciplinary engagement. These developments include GPT-2, GPT-3, and GPT-4, in which ChatGPT's functionality, working mechanism improves considerably (see e.g. Shormani and Alfahad 2025, and references therein).

The second most important trend is *Interval-valued intuitionistic multiplicative linguistic preference relation* (IIMLPR), which is somehow related to the trend *Group decision* (Wu et al. 2018; Zhang et al. 2018; Tang et al. 2019). IIMLPRs were used in relation to *Interval-valued intuitionistic multiplicative linguistic variables* (IVIMLVs) to determine the best methods for decision makers (Kitamura 2023). *Making method* is also to some extent related IIMLPR and *Group decision*. The major citing article is Wu et al. (2018). Another trending theme of AIL research is *Explainable artificial intelligence*. In this trend, Jiménez-Luna et al. (2020) is the top citing article.

Another very crucial trending frontier is *Speaking skill*. This trending issue in AIL research concerns research focusing on the use of AI models, specifically ChatGPT in education. It unveils how human teachers can use ChatGPT in the educational sphere (Fütterer et al. 2023), utilizing it to enhance language acquisition process with reference to speaking skill. In fact, this is one of the major concerns of AI and linguistics specialists, featuring AI models in enhancing classroom activities in speech recognition (Jeon and Lee 2023). An NLP trending issue is manifested by the trending theme *Computing word relatedness*. It is a semantic AI area, where linguistics and AI are intersected (Ben Aouicha et al. 2016). Computing word relatedness is a context-based semantic phenomenon, captured by calculating the meaning of a word depending on the words it co-occurs with. This area of research has sprouted among cognitive linguistics scientists, AI specialists, among others.

The most related research trend to the present study is perhaps *Visual cluster*. Visual clustering is a technique used in data analysis and machine learning to group together similar visual elements or patterns within a dataset, including trending issues and hotspots of a specific realm of human knowledge. This is exactly what we have seen in this study; CiteSpace, a software used to visualize similar articles and group them together. Another area where visualization is employed is for image processing (Jiang et al. 2022).

*Leveraging tweet* constitutes one trending issue in AIL research. Leveraging tweets refers to the strategic use of Twitter and its features to achieve specific goals including increasing brand visibility, driving engagement, or spreading information. Twitter is a popular social media platform that allows users to post and interact with short messages known as tweets. Leveraging tweets can be an effective way for individuals, businesses, and organizations to reach a wide audience and make some impact (Alkhaldi et al. 2022). Alkhaldi et al. (2022) studied leveraging tweets in relation to Covid-19 pandemic, which is characterized with much stress, fear, and psychological problems including depression, hopelessness, loneliness, unknown future, specifically with lack of employment. Their study is a kind of sentiment analysis focusing on a deep learning model known as (SFODLD-SAC), analyzing and classifying COVID-19 tweets, and identifying the sentiments of people during the pandemic.

Additionally, 2018-2024 timeframe comes up with several trending and (re)merging issues and hotspots. The newly trending hotspots include *Novice Programmer, Prioritization, Artificial intelligence,* and *Context-Based Fake News Detection Model*. The reemerging issues include *Making Method, Explainable artificial intelligence*, and *Speaking skill*. These remerging frontiers in AIL research have been discussed above.

However, their reemergence indicates that they are paid much attention to by the academic scholarly.

As for the newly trending research issues, *Natural language processing* retains the first rank. This research frontier did not appear in 1974-2024 timeframe, hence construing a trending hotspot in AIL research in 2018-2024 timeframe. NLP is a well-known area of human endeavor, featuring the relationship between linguistics and AI (McShane and Nirenburg 2021). It was first "born as machine translation, which developed into a high-profile scientific and technological area already in the late 1940s" (McShane and Nirenburg (2021: 22). The second trending issue is *Cross-sectional study*, which is, too, a newly rending issue. Apart from linguistics, there are varied c*ross-sectional* studies, utilizing AI in several spheres including medicine and education (Fütterer et al. 2023; Weidener and Fischer 2024).

Another trending hotspot is *Academic writing*. Academic writing in AI discipline involves the creation of scholarly content related to the theory, research, applications, and advancements in AI. It comprises various forms of academic writing, including research papers, literature reviews, technical reports, conference papers, and journal articles. It could be thought of as the otherwise, i.e. using AI tools such as ChatGPT in academic writing. ChatGPT can generate high-quality academic writing, assist researchers, students, teachers. However, the problem lies in ethical issues, which continues to create hot debate and controversy within the academic community (Amer 2022; Biswas 2023; J Lee 2023; Ortega-Bolaños et al. 2024). Another trending issue in this timeframe is *Using bidirectional encoder representation*. Recall that BERT is a language model, a semantic-based language model, designed to understand the context and meaning of words in a sentence by leveraging a bidirectional approach, which allows the model to consider both the preceding and following words of each word (see also Partee 1995).

The term *Novice programmer* refers to a person who is new to computer programming, typically in the early stages of learning programming, coding and acquiring programming skills (Brinkmann et al. 2023). It constitutes a new trending issue in AIL research for its importance and applicability. Additionally, *Context-based fake news detection model* refers to a language model employing contextual information to identify and classify fake news or misinformation. Context-based approaches consider the surrounding context, such as the content of the news article, its source, and external factors, to make more accurate determinations about the authenticity of the news (Amer et al. 2022).

The second part of the analysis tackles the author's keyword(s) and key contributors to AIL research. The author's keyword(s) analysis provides a vivid picture of where AIL research revolves, the areas where AI scholarly should focus. It also mirrors cluster/trend analysis or DCA, uncovering almost the same trending issues unveiled in ACA. Reconsidering the leading contributors to AIL intellectual knowledge, *IEEE Access, Engineering Application of Artificial Intelligence* and *Artificial Intelligence Review* rank the three top productive journals in this study. As for research centers, *University of Granada*, *King Abdul Aziz University* and *Stanford University* retained the three top ranks in this study. The first is a Spanish university while the second is an Arab university. USA and China are the most leading countries in the world, contributing to AIL research over 51 years.

The AIL landscape, trending issues and hotspots presented in the Tables and Figs. in section 4 clearly demarcate how linguistics "feeds" AI and underlies its development, reflecting a relationship of intersection between linguistics and AI. This intersection finally results in developing powerful language models, specifically with utilizing NNAs. One of these language models, and perhaps the most powerful one, is ChatGPT which can learn deeply once it has been trained on massive amounts. The training data includes books and articles, and all the internet data, and after the training process, it can generate and/or process similar texts (Siu 2023; Sohail et al. 2023). ChatGPT can also perform other tasks such as automated tagging, summarizing, completing codes, bugging, and creating content (Kung et al. 2023; T Lee 2023; Ray 2023; Siu 2023; Shormani 2024c). ChatGPT, or other language models, including automated machine learning models (Eldeeb et al. 2022), come to existence due to AI's long-term goals, which have been to simulate computer to behave like human, with programs designed primarily for processing, generating and/or interpreting human language; language itself is one manifestation of human intelligence. In generative biological approach to the study of language, it is viewed as "a structured and accessible product of the human mind" (Everaert et al. 2015: 729).

To conclude this section, the findings our study has come up with align with several studies. Although our study is considered the first of its kind, there are somehow related studies, which have been conducted on CL/NLP and AI, our study findings align with. For example, the slow increase of AIL publication in the 1980s and 1990s, followed by a significant increase post-2000 (cf. Fig. 1), mirrors trends noted by Church and Liberman (2021) who argue that CL/NLP research has increased gradually, with notable growth in recent decades. Likewise, the surge in publications related to deep learning and neural networks in NLP in our study aligns with trends observed by some researchers (see e.g. Manning 2015). Our findings also emphasize the increasing interdisciplinary collaborations between linguistics and AI, as noted by Arshed et al. (2024). Finally, our study identifies emerging hotspots in neural LLMs such as *Using ChatGPT*, *Novice Programmer, Artificial intelligence,* which align with recent analyses, underscoring the shift towards more advanced AI applications in linguistics as found by Ibarra-Torres and Barona-Pico (2024).

## 6. Conclusion and further implications

To conclude, the correlation between linguistics and AI has been manifested through several and various horizons, the most important of which is deep learning language models including ChatGPT, with its several versions including GPT-1, GPT-2, GPT-3 and GPT-4. This study provides a comprehensive and in-depth scientometric analysis of a well-defined body of AIL research, characterized with rigorous theoretical foundations, specifically from 2018 to 2024. It synthesizes the AIL intellectual knowledge production over 51 years, from 1974 to 2024. It involves 5750 WoSCC articles published in 2123 journals, which are written by 19664 authors belonging to 5580 research centers in 116 countries. The results present several trending issues including *Using ChatGPT*, *Making method, Group Decision, Explainable artificial intelligence, Speaking skill, Computing word relatedness* and *Leveraging tweet* in the 1974-2024 timeframe, they each reflect the correlation between linguistics and AI. The most trending hotspots include *Natural language processing, Novice programmer, Artificial intelligence, Academic writing, BERT* and *Fake news detection model*, which reflect not only the AIL correlation, but also mirror the intellectual knowledge produced, unveiling new horizons, new topics, new applications, and launching powerful deep learning language models including ChatGPT and automated machine learning models (Eldeeb et al. 2022; Baratchi et al. 2024). Thus, the study uncovers the fact that there is a strong correlation between linguistics and AI, best manifested by deep learning language models. Linguistics having profound and powerful machinery apparatus and underlying the scope of psychology, cognition, and biology, and syntax, and semantics, provides AI with the theoretical foundations necessary for programming, training, and working mechanism of language models (Medium 2023; Shormani 2024a/d).

However, this study involves some limitations. One of these limitations concerns the source of the data. Although a reliable and high-quality source, WoS does not encompass all data compared to data from a collection of sources including Scopus, Lens, and PubMed. WoS also involves restricted coverage of many important publications, regional journals, and emerging fields which may be underrepresented or entirely excluded. Additionally, in spite of including all scientometric information in WoS search engine in our study, the missing of author address information and keywords for some WoS-indexed records may influence the results of these studies (see also Liu et al. 2018; Liu 2021; Yeung 2023). This could introduce biases in bibliometric analyses, leading to skewed conclusions about research impact, collaboration networks, and knowledge dissemination, besides "the non-transparency of the data source will hinder the reproducibility of some corresponding studies" (Liu 2019: 1815), and publications of "social sciences and humanities [are] still under-represented" (Asubiaro et al. 2024: 1469, see also Colavizza et al. 2022). These recent studies call for expanding WoSCC 'for the growth of related publications.' Another limitation concerns WoSCC coverage of AIL publications before 1974, because it does not contain articles prior to this date, as was noted in this study. A further limitation has to do with language; all the collected articles are written in English, excluding those in non-English languages, which exhibits language bias (Liu 2017, 2019; Vera-Baceta et al. 2019). Thus, future research could involve data from more than one source such as Scopus, Lens, Google Scholar, to secure full coverage and concretize the actual picture of AIL intellectual landscape, trending issues and hotspots. Further research could also include publications from regional journals, and possibly include more than one language to avoid biases against non-English publications of AIL research.

A very crucial issue that AIL research should address in the future should center around the importance of the ethical issues of deep learning language models such as ChatGPT (Ortega-Bolaños et al. 2024). ChatGPT is widely used by students, teachers, researchers, among others in different and various aspects and for several and various purposes as well. However, this wide range of people, uses and purposes cannot go without rethinking ChatGPT's advantages and disadvantages, ethics and biases, without seeing and determining its harms. Although there has arisen collective consensus of ChatGPT's possible harms, specifically among academics, this awareness has not been produced in a scholarly pool. ChatGPT becomes a reality, a technological necessity, but it should be used in secure conditions, i.e. without affecting its users, their career and even their reputation (J Lee 2023). Thus, there is an urgent need to address this very substantial issue to put the guidelines and boarders of secure future use of LLMs.

Another major challenge faced by linguists and AI specialists is ensuring ethical integrity and mitigating biases in LLMs. AI experts must strive to make these models both ethically responsible and free from bias. For example, ethical considerations require that LLMs deliver fair and unbiased outcomes across various demographic groups to prevent the reinforcement of societal disparities (Mutashar 2024; Shormani 2024d). Addressing these issues is crucial in ensuring equitable treatment and reducing unintended discrimination. Additionally, LLMs are often perceived as "black boxes," making their decision-making processes difficult to interpret. To foster trust and transparency, AI specialists should prioritize model interpretability, allowing users to verify and understand how information is processed. Another critical ethical concern involves privacy and data protection. Since LLMs rely on vast amounts of internet data for training, often including sensitive personal information, there is a risk of violating privacy rights. To overcome this, AI specialists must adhere to privacy laws and implement safeguards to minimize data misuse and leakage (Shormani 2024d).

An actionable strategy that linguists and AI experts should focus on is AI biases in LLMs. This aspect can lead to harmful consequences, particularly when AI tools reinforce existing societal inequalities. Biases can be of various types including gender, ethnicity, and linguistics. As for gender, a recruitment algorithm, for example, trained predominantly on male-dominated job applications may favor male candidates, highlighting how algorithmic design can introduce bias when certain groups are underrepresented (Lion et al. 2024; Shormani 2024d). Similarly, LLMs that integrate user feedback risk amplifying biases inherent in user behavior (Formiga et al. 2015). Ensuring equal representation in LLMs is essential, as in the case of disabled people or ethnic communities who might be underrepresented or misrepresented in training data, leading to biased outputs. Finally, linguistic bias is another significant challenge; many LLMs are primarily trained on English data, neglecting languages such as Arabic, Persian, and Urdu, or indigenous languages, which may lead to bias against these languages and non-English languages, in general (Shormani 2024d). Thus, AIL scholars should address these areas in their future studies.

A final practical recommendation concerns LLMs, specifically ChatGPT. To ensure the responsible and effective output of ChatGPT, AIL future research should prioritize ethical guidelines that address its potential biases, privacy concerns, and lack of transparency. As ChatGPT is increasingly used by students, educators, and researchers, it is essential to establish clear boundaries for its application to prevent misuse and protect users' professional integrity (Shormani 2024d, Shormani and Alfahad 2025). Researchers should also work on reducing biases related to gender, ethnicity, disability, and language by improving training data diversity and ensuring equitable representation. Additionally, enhancing the interpretability of ChatGPT's output can help build trust and understanding of its decision-making processes. Expanding multilingual capabilities by integrating non-English languages, specifically "low-source languages such as Hindi, Bengali, Amharic, and Irish. Due to the poor linguistic data these languages have, models developed for them need further research" (Shormani 2024d: 16-17). This aspect is crucial to minimizing these ChatGPT's linguistic biases and fostering inclusivity.

**Declaration of Interest**

There are no competing interests to declare.

**Data availability statement**

The dataset underlying the results of this study will be available from the author upon a reasonable request to the editor.

## References

Ackema P, Brandt P, Schoorlemmer M, Weerman F (2006) The role of agreement in the expression of arguments. Arguments and Agreement. 1-34.

Alkhaldi N A, Asiri Y, Mashraqi A M, Halawani H T, Abdel-Khalek S, Mansour R F (2022) Leveraging tweets for artificial intelligence driven sentiment analysis on the COVID-19 pandemic. In Healthcare, 10(5), 910). MDPI.

AlSagri HS, Farhat F, Sohail SS, Saudagar AK. ChatGPT or Gemini (2024) Who makes the better scientific writing assistant? Journal of Academic Ethics. 31:1-5.

Alsagri HS, Sohail SS (2024) Fractal-inspired sentiment analysis: Evaluation of large language models and deep learning methods. Fractals. 18;2440056:12.

Amer E, Kwak K, El-Sappagh S (2022) Context-based fake news detection model relying on deep learning models. Electronics. 11:1255. https://doi.org/10.3390/electronics11081255.

Anas M, Saiyeda A, Sohail SS, Cambria E, Hussain A (2024) Can generative AI models extract deeper sentiments as compared to traditional deep learning algorithms?. IEEE Intelligent Systems. 30;39(2):5-10.

Anwar K, Zafar A, Iqbal A, Sohail SS, Hussain A, Karaca Y, Hijji M, Saudagar AK, Muhammad K (2023) Artificial intelligence-driven approach to identify and recommend the winner in a tied event in sports surveillance. Fractals. 28;31(10):2340149.

Arshed MA, Gherghina ȘC, Dewi C, Iqbal A, Mumtaz S (2024) Unveiling AI-generated financial text: A computational approach using natural language processing and generative artificial intelligence. Computation. 15;12(5):101.

Asubiaro T, Onaolapo S, Mills D (2024) Regional disparities in Web of Science and Scopus journal coverage. Scientometrics. 129(3), 1469-1491. https://doi.org/10.1007/s11192-024-04948-x

Ballouk H, Jabeur S, Challita S, Chen C (2024) Financial Stability: A scientometric analysis and research agenda. Res. Intern. Bus. Fin. 70: 102294: 1-15 http://dx.doi.org/10.1016/j.ribaf.2024.102294

Baratchi, M., Wang, C., Limmer, S. et al. (2024) Automated machine learning: past, present and future. Artif Intell Rev 57, 122. https://doi.org/10.1007/s10462-024-10726-1

Baroni M, Lenci A (2010) Distributional memory: a general framework for corpus-based semantics Comput. Linguist. 36: 673–721.

Ben Aouicha M, Hadj Taieb M, Ben Hamadou A (2016) LWCR: Multi-layered wikipedia representation for computing word relatedness. Neurocomputing. 216:816-843. http://dx.doi.org/10.1016/j.neucom.2016.08.045

Biswas S (2023). ChatGPT and the future of medical writing. Radiology. 307(2), 223312. https://doi.org/10.1148/RADIOL.223312

Brinkmann L, Baumann F, Bonnefon J et al. (2023) Machine culture. Nat. Hum. Behav. **7:**1855–1868. https://doi.org/10.1038/s41562-023-01742-2.

Chen C (2003) Mapping scientific frontiers: the quest for knowledge visualization. Springer, London.

Chen C (2006) CiteSpace II: Detecting and visualizing emerging trends and transient patterns in scientific literature. Journal of the American Society for Information Science and Technology. 573:359-377.http://dx.doi.org/10.1002/asi.20317

Chen C (2017) Science Mapping: A Systematic Review of the Literature. J. Data Info. Sci. 2(2):1-40 http://dx.doi.org/10.1515/Jdis-2017-0006
Chen C, Song M (2019) Visualizing a field of research: A methodology of systematic scientometric reviews. PloS one. 31;14(10):e0223994. https://doi.org/10.1371/journal.pone.0223994
Colavizza G, Peroni S, Romanello M (2022) The case for the Humanities Citation Index (HuCI): A citation index by the humanities, for the humanities. International Journal on Digital Libraries. https://doi.org/10.1007/s00799-022-00327-0
Dergaa I, Chamari K, Zmijewski P, Ben Saad H (2023) From human writing to artificial intelligence generated text: examining the prospects and potential threats of ChatGPT in academic writing. Biol Sport. 40(2):615–622.http://dx.doi.org/10.5114/biolsport.2023.125623
Devlin J, Chang MW, Lee K, Toutanova K (2019). Bert: Pre-training of deep bidirectional transformers for language understanding. In: Proceedings of the 2019 conference of the North American chapter of the association for computational linguistics: human language technologies, volume 1 (long and short papers). 4171-4186.
Eldeeb H, Maher M, Elshawi R, Sakr S (2022) AutoMLBench: a comprehensive experimental evaluation of automated machine learning frameworks. Expert Systems with Applications. 1;243:122877.
Ernst N A, Bavota G (2022) Ai-driven development is here: Should you worry?. *IEEE Software*, *39*(2), 106-110.
Ettinger A (2020) What BERT is not: lessons from a new suite of psycholinguistic diagnostics for language models.Trans. Assoc. Comput. Linguist. 8: 34–48.
Ettinger A, Elgohary A, Phillips C, Resnik P (2018) Assessing composition in sentence vector representations. Proceedings of the 27th International Conference on Computational Linguistics.1790–1801.
Everaert M, Huybregts M, Chomsky N, Berwick R, Bolhuis J (2015) Structures, not strings: linguistics as part of the cognitive sciences. Trends Cogn. Sci. 19: 729–743.
Fodor J, Pylyshyn Z (1988) Connectionism and cognitive architecture: a critical analysis. Cognition. 28(1):23-71.
Formiga L, Barrón-Cedeno A, et al. (2015) Leveraging online user feedback to improve statistical machine translation. Journal of Artificial Intelligence Research. 54, 159-192. https://doi.org/10.1613/jair.4716
Fütterer T, Fischer C, Alekseeva A et al. (2023) ChatGPT in education: global reactions to AI innovations. Sci Rep 13, 15310(2023). https://doi.org/10.1038/s41598-023-42227-6
Goodfellow I, Bengio Y, Courville A, Bengio Y (2016) *Deep learning*. Cambridge: MIT Press.
Gulordava K, Bojanowski P, Grave E, Linzen T, Baroni M (2018) Colorless green recurrent networks dream hierarchically. Proceedings of the 2018 Conference of the North American Chapter of the Association for Computational Linguistics: Human Language Technologies, pp. 1195–1205.
Howard J (2019) Artificial intelligence: Implications for the future of work. Amer. J. indus. Med. 62(11): 917–926. https://doi.org/10.1002/ajim.23037
Ibarra-Torres F, Barona-Pico V (2024) Use ChatGPT to inspect software developed by novice programmers. In: Garcia MV, Gordón-Gallegos C, Salazar-Ramírez A, Nuñez C (eds) *Proceedings of the International Conference on Computer Science, Electronics and Industrial Engineering (CSEI 2023)*. Lecture Notes in Networks and Systems, 797. Springer, Cham. https://doi.org/10.1007/978-3-031-70981-4_8

Jeon J, Lee S (2023) Large language models in education: A focus on the complementary relationship between human teachers and ChatGPT. Educ. Inf. Technol. 28:15873-15892. https://doi.org/10.1007/s10639-023-11834-1

Jiang W, Zhu M, Fang Y, Shi G, Zhao X, Liu Y (2022) Visual cluster grounding for image captioning, in IEEE Transactions on Image Processing. 31:3920-3934. http://dx.doi.org/doi: 10.1109/TIP.2022.3177318

Jiao W, Wenxuan W, Huang J, et al. (2023) Is ChatGPT a good translator? Yes with GPT-4 as the engine. https://arxiv.org/abs/2301.08745Z.

Jiménez-Luna J, Grisoni F, Schneider G (2020) Drug discovery with explainable artificial intelligence. Nat Mach Intell. 2:573–584. https://doi.org/10.1038/s42256-020-00236-4

Kenny D (2022) Human and machine translation. In: Kenny D (ed.), Machine translation for everyone: Empowering users in the age of artificial intelligence. Language Science Press, Berlin. 23–49.

Kitamura, F (2023) ChatGPT is shaping the future of medical writing but still requires human judgment. Radiology. 307:230171. https://doi.org/10.1148/RADIOL.230171

Koeneman O, Zeijlstra H (2014) The rich agreement hypothesis rehabilitated. Linguistic Inquiry. 45(4), 571-615.

Kolides A, Nawaz A, Rathor A, Beeman D, Hashmi M. Fatima S, Jararweh Y (2023) Artificial intelligence foundation and pre-trained models: fundamentals,applications, opportunities, and social impacts. Simul. Model. Pract. Theory 126: 02754. https://doi.org/10.1016/j.simpat.2023.102754.

Kung T, Morgan C, Arielle M et al. (2023) Performance of ChatGPT on USMLE: Potential for AI-assisted medical education using large language models. PloS Dig. Heal. 1-12. https://doi.org/10.1371/journal.pdig.0000198

Lee T (2023) Artificial intelligence and posthumanist translation: ChatGPT vs the translator. Appl. Ling. Rev.doi: 10.1515/applirev-2023-0122

Lee, J. (2023) Can an artificial intelligence chatbot be the author of a scholarly article? J. Educ. Eval. Heal. Profess. 20(6). https://doi.org/10.3352/JEEHP.2023.20.6

Li B Z, Nye M, Andreas J (2021) Implicit representations of meaning in neural language models. arXiv preprint arXiv:2106.00737.

Li D, Liu Y, Huang J Wang Z (2023) A trustworthy view on explainable artificial intelligence method evaluation, Computer. 56:50-60. https://doi.org/10.1109/MC.2022.3233806

Liao H, Xu Z, Herrera-Viedma E et al. (2018) Hesitant fuzzy linguistic term set and its application in decision making: A state-of-the-art survey. Int. J. Fuzzy Syst. 20: 2084–2110.

Linzen T, Baroni M (2021) Syntactic structure from deep learning. Ann. Rev. Ling. 7:195-212. https://doi.org/10.1146/annurev-linguistics-032020-051035

Linzen T, Dupoux E, Goldberg Y (2016) Assessing the ability of LSTMs to learn syntax-sensitive dependencies. Trans. Assoc. Comput. Linguist. 4:521–535.

Lion K C, Lin Y H, Kim, T. (2024) Artificial intelligence for language translation: The equity is in the details. JAMA, 332(17), 1427-1428.

Liu W (2017) The changing role of non-English papers in scholarly communication: Evidence from Web of Science's three journal citation indexes. *Learned Publishing*, *30*(2), 115-123.

Liu W (2021) Caveats for the use of Web of Science Core Collection in old literature retrieval and historical bibliometric analysis. *Technological Forecasting and Social Change*, *172*, 121023.

Liu W, Hu G, Tang L (2018) Missing author address information in Web of Science—An explorative study. *Journal of Informetrics*, *12*(3), 985-997.

Liu W, Ni R, Hu G (2024) Web of science core collection's coverage expansion: The forgotten arts & humanities citation index. *Scientometrics*, *129*(2), 933-955.

Liu, W. (2019). The data source of this study is Web of Science Core Collection? Not enough. Scientometrics. 121(3), 1815-1824. https://doi.org/10.1007/s11192-019-03238-1

Manning CD (2015) Computational linguistics and deep learning. *Computational Linguistics*. 1;41(4):701-7.

Marelli M, Baroni M (2015) Affixation in semantic space: modeling morpheme meanings with compositional distributional semantics. Psychol. Rev. 122:485–515.

Maruthi S, Dodda SB, Yellu RR, Thuniki P, Reddy SR (2021) Deconstructing the Semantics of Human-Centric AI: A Linguistic Analysis. Journal of Artificial Intelligence Research and Applications. 1(1):11-30.

McShane M, Nirenburg S (2021) *Linguistics for the Age of AI*. MIT Press.

Medium (2023). Linguistics, language models, and artificial intelligence. [Online], available at: https://medium.com/@hmsajjad/on-the-intersection-of-linguistics-language-models-and-artificial-intelligence-an-overview-d3c89c67690d. Accessed Jan. 3, 2024.

Minsky, M. (*1961*). Steps toward artificial intelligence. In *Proceedings of the IRE* 49: 8-30.

Mutashar M K (2024) Navigating ethics in AI-driven translation for a human-centric future. Academia Open, 9(2), 10-21070.

Nadeem M, Sohail SS, Javed L, Anwer F, Saudagar AK, Muhammad K (2024) Vision-enabled large language and deep learning models for image-based emotion recognition. Cognitive Computation.16(5):2566-79.

Ortega-Bolaños R, Bernal-Salcedo J, Germán Ortiz M, et al. (2024) Applying the ethics of AI: a systematic review of tools for developing and assessing AI-based systems. Artif Intell Rev 57, 110. https://doi.org/10.1007/s10462-024-10740-3

Partee B (1995) Lexical semantics and compositionality. In: Gleitman LR, Liberman M (eds.), Language: An invitation to cognitive science. The MIT Press, pp. 311-360).

Pavlick E (2022) Semantic structure in deep learning. Ann. Rev. Ling. 8:447-471. https://doi.org/10.1146/annurev-linguistics-031120-122924

Polinsky M, Potsdam E (2001) Long-distance agreement and topic in Tsez. Natural Language & Linguistic Theory, 19(3), 583-646.

Porcel C, Ching-López A, Lefranc G, Loia V, Herrera-Viedma E (2018) Sharing notes: An academic social network based on a personalized fuzzy linguistic recommender system. Eng. Applic. Artif. Intel. 75:1-10. https://doi.org/10.1016/j.engappai.2018.07.007.

Qamar MT, Yasmeen J, Pathak SK, Sohail SS, Madsen DØ, Rangarajan M (2024) Big claims, low outcomes: fact checking ChatGPT's efficacy in handling linguistic creativity and ambiguity. Cogent arts & humanities. 31;11(1):2353984.

Rasmy L, Xiang Y, Xie Z, Tao C, Zhi D (2021) Med-BERT: pretrained contextualized embeddings on large-scale structured electronic health records for disease prediction. NPJ Digit. Med. 4(86) https://doi.org/10.1038/s41746-021-00455-y.

Ray P P (2023) Background, applications, key challenges, bias, ethics, limitations and future scope. Internet of Things and Cyber-Physical Systems 121-154.https://doi.org/10.1016/j.iotcps.2023.04.003

Ribeiro MT, Singh S, Guestrin C (2016) Model-agnostic interpretability of machine learning. arXiv preprint arXiv:1606.05386. 2016 Jun 16.
Rouveret A (2008) Phasal agreement and reconstruction In: Freidin R, Otero C, Zubizarreta M (eds.), *Foundational issues in linguistic theory*. Cambridge, MA: MIT Press, pp. 167-196.
Shormani M Q (2014a) The nature of language acquisition: Where L1 and L2 acquisition meet? Journal of Literature, Languages and Linguistic, 4. 24-34.
Shormani M Q (2017) SVO, (silent) topics and the interpretation of referential pro: A discourse-syntax interface approach. *Italian Journal of Linguistics*, *29*(2), 91-159.
Shormani MQ (2013). *An introduction to English syntax: A generative approach*. Lambert Academic Publishing, Germany.
Shormani MQ (2014b) Collocability difficulty: A UG-based model for stable acquisition. Journal of Literature, Languages and Linguistics. 4, 54-64.
Shormani MQ (2016) Biolinguistics, the 'magnetic' mechanism of Language Faculty and language acquisition. Journal of Teaching and Teacher Education, 4(01), 71-88.
Shormani MQ (2023) L2 acquisition of Wh-interrogatives at the syntax-discourse interface: interface hypothesis again. F1000Research*,* 12.
Shormani MQ (2024a) Linguistics contribution to artificial intelligence. (In press).
Shormani MQ (2024b) *Introducing minimalism: A parametric variation*. Lincom Europa Press.
Shormani MQ (2024c). Can ChatGPT capture swearing nuances? Evidence from translating Arabic oaths. (To appear, *JPSC*).
Shormani MQ (2024d) Artificial intelligence contribution to translation industry: looking back and forward. arXiv preprint arXiv:2411.19855. 2024 Nov 29.
Shormani MQ (2025) Non-native speakers of English or ChatGPT: Who thinks better? F1000Research. https://doi.org/10.12688/f1000research.161306.1
Shormani MQ, Alfahad A (2025) Artificial Intelligence or Human: The use of ChatGPT in the academic translation for religious texts. (To appear in Sage Open).
Shormani MQ, Qarabesh M (2018) Vocatives: correlating the syntax and discourse at the interface. Cogent Arts & Humanities, 5(1), 1469388.
Shwartz V, Dagan I (2019) Still a pain in the neck: evaluating text representations on lexical composition. Trans. Assoc. Comput. Linguist. 7:403–419.
Siu S C (2023) ChatGPT and GPT-4 for professional translators: exploring the potential of large language models in translation. Preprint. 1-36. Available from https://papers.ssrn.com/sol3/papers.cfm?abstract_id=4448091
Sobel I (1974) On calibrating computer controlled cameras for perceiving 3-D scenes. *Artificial intelligence*, *5*(2), 185-198.
Sohail SS, Farhat F, Himeur Y, Nadeem M, Madsen D, Singh Y, Atalla S, Mansoor W (2023) Decoding ChatGPT: A taxonomy of existing research, current challenges, and possible future directions. Journal of King Saud University-Computer and Information Sciences. 1;35(8):101675. https://doi.org/10.1016/j.jksuci.2023.101675

Sohail SS, Siddiqui J, Ali R (2018) Book recommender system using fuzzy linguistic quantifiers. In *Applications of Soft Computing for the Web,* 47-60. Springer.
Tang H, Shi Y, Dong P (2019) Public blockchain evaluation using entropy and TOPSIS. Expert Syst. Appl. 117: 204-210.

Thrush T, Wilcox F, Levy R (2020) Investigating novel verb learning in BERT: selectional preference classes and alternation-based syntactic generalization. arXivpreprint arXiv:2011.02417.

Turing AM (1950) Computing machinery and intelligence. Mind, New Series 59(236): 433-460.1950.

Van Dis E A, Bollen J, Zuidema W, Van Rooij R, and Bockting C L (2023) ChatGPT: five priorities for research. *Nature*, *614*(7947), 224-226.

Vaswani, A. (2017). Attention is all you need. *Advances in Neural Information Processing Systems.*

Vera-Baceta MA, Thelwall M, Kousha K (2019) Web of Science and Scopus language coverage. *Scientometrics*, *121*(3), 1803-1813.

Wang EH, Wen CX. Building (2025) A Unified AI-centric Language System: analysis, framework and future work. arXiv preprint arXiv:2502.04488.

Weidener L, Fischer M (2024) Artificial intelligence in medicine: Cross-sectional study among medical students on application, education, and ethical aspects. JMIR Medical Education. 10:e51247. https://doi.org/10.2196/51247

Winograd T (1971) Procedures as a representation for data in a computer program for understanding natural language. Ph.D. Dissertation, MIT.

Wu X, Liao H, Xu Z et al. (2018) Probabilistic linguistic MULTIMOORA: A multicriteria decision making method based on the probabilistic linguistic expectation function and the improved Borda rule. IEEE Transactions on Fuzzy Systems. 26(6): 3688-3702.

Yeung A W K, (2023) A revisit to the specification of sub-datasets and corresponding coverage timespans when using Web of Science Core Collection. *Heliyon*, *9*(11).

Zhang Y, Xu Z, Liao H (2018) An ordinal consistency-based group decision making process with probabilistic linguistic preference relation. *Information Sciences*, *467*, 179-198.